\documentclass[10pt, a4paper, onecolumn]{article}

\usepackage{arxiv}
\usepackage[toc]{appendix}
\usepackage[utf8]{inputenc} % allow utf-8 input
\usepackage[T1]{fontenc}    % use 8-bit T1 fonts
\usepackage{hyperref}       % hyperlinks
\usepackage{url}            % simple URL typesetting
\usepackage{booktabs}       % professional-quality tables
\usepackage{amsfonts}       % blackboard math symbols
\usepackage{nicefrac}       % compact symbols for 1/2, etc.
\usepackage{microtype}      % microtypography
\usepackage{lipsum}
\usepackage{amsmath}
\usepackage{cite}
\usepackage{amssymb}
\usepackage{verbatim}
\usepackage{enumerate}
\usepackage{mdwlist} % for compact lists
\usepackage{color}

\usepackage{algpseudocode}
\usepackage{algorithmicx}
\usepackage{lineno}
\usepackage{framed,multirow}
\usepackage[final]{pdfpages}
\usepackage{latexsym}
\usepackage{wrapfig}

\usepackage{url}
\usepackage[table]{xcolor}
\usepackage{xcolor}
\usepackage{graphicx}
\usepackage{subfigure}
\usepackage{verbatim}
\usepackage{bm}
\usepackage{authblk}
\usepackage{tikz}
\usepackage{amsmath}
\usepackage[linesnumbered,ruled]{algorithm2e}
\usepackage[space]{grffile}
\usetikzlibrary{shapes,arrows}

\title{Improving Deep Neural Network Random Initialization Through Neuronal Rewiring}

\author[1,2]{Leonardo Scabini}
\author[2]{Bernard De Baets}
\author[1]{Odemir M. Bruno}
\affil[1]{\small{S\~{a}o Carlos Institute of Physics, University of S\~{a}o Paulo, São Carlos - SP, PO Box 369, 13560-970, Brazil (email: $\{$scabini,bruno$\}$@ifsc.usp.br)}}
\affil[2]{\small{KERMIT, Department of Data Analysis and Mathematical Modelling, Ghent University, Coupure links 653, postal code 9000, Ghent, Belgium (email: $\{$Leonardo.Scabini,Bernard.DeBaets$\}$@UGent.be)}}

\begin{document}
\maketitle

\begin{abstract}
The deep learning literature is continuously updated with new architectures and training techniques. However, weight initialization is overlooked by most recent research, despite some intriguing findings regarding random weights. On the other hand, recent works have been approaching Network Science to understand the structure and dynamics of Artificial Neural Networks (ANNs) after training. Therefore, in this work, we analyze the centrality of neurons in randomly initialized networks. We show that a higher neuronal strength variance may decrease performance, while a lower neuronal strength variance usually improves it. A new method is then proposed to rewire neuronal connections according to a preferential attachment (PA) rule based on their strength, which significantly reduces the strength variance of layers initialized by common methods. In this sense, PA rewiring only reorganizes connections, while preserving the magnitude and distribution of the weights. We show through an extensive statistical analysis in image classification that performance is improved in most cases, both during training and testing, when using both simple and complex architectures and learning schedules. Our results show that, aside from the magnitude, the organization of the weights is also relevant for better initialization of deep ANNs.
\end{abstract}

% keywords can be removed

\section{Introduction}
Over the past decade, Artificial Neural Networks (ANNs) have been dominating Artificial Intelligence
% in means at one specific moment in that decade
applications in various domains. This diffusion mainly came after advances in deep learning, a collection of techniques for training large ANNs to solve complex tasks through several levels of abstraction. The successful training of deep models is usually possible by careful architecture design, high model complexity, and large-scale datasets. Although ANNs are broadly applied and a lot of research has been done to describe their functioning, understanding the broad impacts of their degrees of freedom is still a challenge since modern deep models hold millions or even billions of parameters. 
%Such models with so many degrees of freedom may exhibit unexpected behavior~\cite{nguyen2015deep}, for instance, the sensitivity to adversarial samples~\cite{goodfellow2014explaining}.

Recent works~\cite{picard2021torch,wightman2021resnet} discuss the impacts of stochastic and random hyperparameters (different random seeds) used during the construction and training of deep ANNs, and also the expected uncertainty in this process~\cite{huang2021quantifying}. One important finding is that although the performance variance caused by different seeds can be relatively small, outliers are easily found, \emph{i.e.}, models with a performance much above or below the average. However, the element that poses the highest degree of freedom in this context (at least in theory), the distribution of initial random weights, is overlooked by most of the current research.

Some intriguing properties have also been observed, for instance, specific subsets of random weights that make training of sparse ANNs particularly effective~\cite{frankle2018lottery}. More surprisingly, these random weights may not even require additional training~\cite{zhou2019deconstructing,ramanujan2020s}. It has also been shown that successfully trained ANNs usually converge to a neighborhood of weights close to their initial configuration~\cite{li2018learning,jesus2021effect}. These works corroborate the importance of initial weights and also point toward the existence of particular random structures related to better initial models. Nevertheless, most works initialize ANNs with simple methods that only define bounds for random weight sampling. Moreover, the effects caused by the randomness of these methods are most of the time ignored, \emph{i.e.}, researchers arbitrarily choose to consider either a single random trial or to select the one that yields the better performance within a set of trials, which is a way to over-tune and overestimate the model capabilities.

The randomness involved in ANN initialization raises some questions. For instance, what are the structural properties among a large set of randomly initialized networks? Is there any meaningful pattern of random weights that could help to improve initial models, making the task less trial-and-error based? One possible alternative to approach large and complex ANNs is through Network Science (or Complex Networks), a prominent research field bringing together graph theory and statistical physics to study real-world network-like complex systems such as brain dynamics, social networks, protein interaction, and many others~\cite{costa2011analyzing}. Few works have analyzed ANNs from the point of view of Network Science, and they usually do not consider a reliable sample size to account for the variance caused by the huge number of possible states of random initial weights. %Moreover, these works usually focus on ANN models after training.

In this work, we study the structural properties of random ANNs through the lens of Network Science. We focus on the most common weight initialization approaches, which rely solely on the information of a layer to produce its random weights. This is the most commonly available approach in widely used libraries, such as Pytorch and TensorFlow, and is also employed by most of the current research. Our main contributions are:

\begin{itemize}
    \item[(i)] We describe the neuronal strength of randomly initialized ANNs and its correlation with performance.
    
    \item[(ii)] A new method is proposed, which consists of gradually rewiring weights of individual neurons through preferential attachment (PA). This process considerably reduces the strength variance, resulting in a more homogeneous neural distribution while keeping the same weight distribution. In other words, these properties are achieved by simply reorganizing previously initialized weights. The approach can be employed in virtually any ANN architecture and depth coupled with any common initializer. Since we only modify the initial weight matrices, it also works in conjunction with any technique for data and architecture regularization, learning schedule, etc. We provide an extensive statistical evaluation of how this approach affects the model behavior during and after training. 
    
    \item[(iii)] Our findings bring together concepts from Network Science and ANNs, showing that it is a promising approach to analyze and improve weight initialization, and also corroborating previous works regarding the structure and functioning of ANNs.
\end{itemize}

All our code for the proposed PA rewiring method and the experimental evaluation is made publicly available\footnote{\url{www.github.com/scabini/PArewiring_weights}}. The paper is divided into the following sections: Section~\ref{sec:2} provides the fundamental background on ANNs and Network Science; Section~\ref{sec:method} describes our proposal for ANN random weight initialization; Section~\ref{sec:results} contains the experimental results and discussions, while Section~\ref{sec:conclusion} draws the most important conclusions from this work.

\section{Background}\label{sec:2}

For supervised classification of images, an ANN can be represented as a function $f(x,\theta_f)=y$, where $x$ is the input image, $y$ is the class label assigned, and $\theta$ the model parameters. There is a wide range of approaches to building the~function $f$~\cite{khan2020survey,han2022survey}, referred to as the model architecture. Here, we focus on the model parameters $\theta= 
\left( W_{n_l, n_{l+1}}\right)_l$, a collection of weight matrices each representing the connections between two consecutive layers, where $l$ represents layer indices from the input to the output layer, with size $n_l$. When training a model from scratch, these parameters are usually randomly initialized and then optimized using gradient descent techniques to reduce the error on the task at hand. A common approach, known as transfer learning, consists of starting the training procedure 
%of $f(x,\theta_g)$ 
with a set of parameters $\theta_g$ optimized in another similar task.
%$g(x_g,\theta_g)=y_g$. 
However, our focus here is on the random initialization of $\theta_f$ for training from scratch, \emph{i.e.}, when no additional information or previous knowledge is available, or when a new architecture is developed.

% \subsection{Artificial Neural Networks}

%Deep Learning started to flourish after the ILSVRC \cite{russakovsky2015imagenet} 2012 results, when the winner, a deep convolutional network (AlexNet) \cite{krizhevsky2012imagenet}, surpassed all other competitors by a significant margin. AlexNet represents the first successful development of a deep ANN, both in terms of theory and practical implementation, that was able to train effectively in a large-scale image database.

%This achievement caused machine learning research to be dominated by deep ANNs, from computer vision to natural language processing \textbf{CITAR}, time series prediction \textbf{CITAR}, and more.

% \subsubsection{Activation Functions}

% Rectifier Linear Unit (RELU) $f(x) = x^+ = max(0,x)$ \cite{glorot2011deep}
% Parametric RELU (PRELU) \cite{he2015delving}
% Gaussian error linear unit (GELU) \cite{hendrycks2016gaussian}
% scaled exponential linear unit (SELU) \cite{klambauer2017self}

\subsection{Weight Initialization for Deep Neural Networks}

%Deep Learning started to flourish after the ILSVRC \cite{russakovsky2015imagenet} 2012 results, when the winner, a deep convolutional network (AlexNet) \cite{krizhevsky2012imagenet}, surpassed all other competitors by a significant margin. AlexNet represents the first successful development of a deep ANN, both in terms of theory and practical implementation, that was able to train effectively in a large-scale image database. %This achievement caused machine learning research to be dominated by deep ANNs, from computer vision to natural language processing \textbf{CITAR}, time series prediction \textbf{CITAR}, and more.

Initialization of ANN weights is an old topic, and here we focus on the initialization of deep models, a trend that became popular after the beginning of this century. The reader may also refer to~\cite{narkhede2021review} for a broader review on weight initialization techniques. Early works on deep ANN weight initialization~\cite{hinton2006reducing,bengio2007greedy} focused on unsupervised pretraining layers, which could improve training speed and generalization performance of standard gradient descent. Later on, Glorot and Bengio~\cite{glorot2010understanding} proposed to sample weights from a uniform distribution centered on zero with bounds defined by the activation variance of linear neurons. This technique
% B: strange, his name is Xavier Glorot, so they either use first name or family name, why is Bengio left out? see also 255; L: Is common to see this method referenced only as "Glorot initialization" or "Xavier initialization", but let us refer to both authors
 defines bounds for either uniform ($\pm\sqrt{\frac{6}{n_l+n_{l+1}}}$) or normal ($\sigma=\sqrt{\frac{2}{n_l+n_{l+1}}}$) weight distributions. However, their derivation defines just the bounds of values for randomly sampling weights, and it also does not account for most recent non-linear activation functions such as rectified linear units (ReLU)~\cite{glorot2011deep}.

After the deep learning research diffusion, many methods focused on more specific techniques for successfully training deeper models. He et al.~\cite{he2015delving} (Kaiming initializer) extended the Glorot and Bengio~\cite{glorot2010understanding} method for ReLU nonlinearities, which consists of varying the bounds for random weight sampling to account for the ReLU variance ($\pm\sqrt{\frac{6}{n_l}}$ for uniform or $\sigma=\sqrt{\frac{2}{n_l}}$ for normal distributions).
%This method is sometimes also referred to as "Microsoft weight initialization" \cite{klambauer2017self}.
Saxe et al.~\cite{saxe2013exact} suggest that the weight matrix should be chosen as a random orthogonal matrix, and show that this has similar effects as pretraining layers, and helps to train deep models faster and better. 
%They achieve this using a mathematical condition for error backpropagation, namely dynamical isometry, where deep random orthogonal linear networks achieve perfect dynamical isometry and achieve depth independent learning times.
This approach was further explored more recently~\cite{hu2020provable}, showing that in deep networks the width needed for efficient convergence with orthogonal initialization is independent of the depth, whereas the width needed for efficient convergence with Gaussian initialization scales linearly with the depth. Another approach considers random walk initialization~\cite{sussillo2014random}, which keeps constant the log of the norms of the backpropagated errors, and was shown to make it possible to train networks with depths of up to 200 layers. They achieve a random walk analogy by deriving the estimated backpropagated normalized errors flowing through random layers in a deep network. Hendrycks and Gimpel~\cite{hendrycks2016adjusting}  
propose to scale the random weight bounds by accounting for the variance when using dropout layers. It was also shown that the Hessian matrix can be considered for ANN initialization~\cite{skorski2021revisiting}, e.g., using an approximation of the layers’ Hessians via Jacobian products and the chain rule.

%Another approach toward ANN initialization is the re-parametrization of random weights. In \cite{mishkin2015all} the authors propose a technique named Layer-sequential unit-variance initialization (LSUV), which starts with orthonormal-initialized weight matrices \cite{saxe2013exact} and then normalizes the variance of the layer output to be one, given training batches. \cite{salimans2016weight} proposes a re-parametrization of the weight vectors by decoupling their length from their direction using statistics computed from a batch of data (weight normalization). However, these techniques depend on the training data and may be sensitive when estimating the initial values. %Moreover, these techniques are usually also applied throughout training, and not only at initialization, introducing additional learning costs.

A recent approach to deep ANN pruning, called the lottery ticket hypothesis (LTH)~\cite{frankle2018lottery}, suggests that there exist random subnetworks ("winning lottery tickets") that can achieve similar or better performance than the whole model in the same number of training steps. 
%These subnetworks can be found with a method called iterative magnitude pruning (IMP) after the model is trained.
After pruning, the remaining parameters can be reset to their random initial state and the subnetwork can be trained again to a similar optimum compared to the original larger network. Therefore, this subnetwork is a subset of the random initial weights and is not related to some organization that emerges during training. Follow-up works on the LTH~\cite{zhou2019deconstructing,ramanujan2020s,frankle2020early} propose that random weights that "win the lottery" are highly correlated with each other, and may not even require training to achieve comparable performance to that of the trained models. 
%These effects also resemble Randomized neural networks \cite{schmidt1992feedforward,pao1992functional,huang2006extreme} (sometimes also referred to as Extreme Learning Machines), a class of models composed of hidden layers with random weights where only the output layer is trained.
The results in~\cite{jesus2021effect} show that the successful training of ANNs via Stochastic Gradient Descent (SGD)~\cite{sutskever2013importance} converges to the close neighborhood of the initial configuration of its weights. These findings corroborate that specific random initializations may have great effects on training dynamics and model performance. 

%For instance, the fact that bigger networks contain sub-networks with compared performance suggests that increasing the search space, i.e., the size of the initial random "guess", increases the chances of also finding smaller "good" solutions inside it.

%A common trend among these previous works is that they agree that correctly scaled random matrices help the initialization of deep networks. 

\subsection{Network Science}

Network Science, also referred to as Complex Networks, is a research field combining graph theory and statistical physics (complex systems) arising from the need to analyze the structure and dynamics of real-world networks. It usually consists of describing heterogeneous, temporal, and adaptive patterns of interactions in such networks. This methodology is employed for the modeling and characterization of natural phenomena in a broad range of areas~\cite{costa2011analyzing}. For instance, these concepts were broadly employed recently to model and understand the COVID-19 pandemic~\cite{scabini2021social}. 

A network can be defined by a tuple $G=(V, E)$, where $V=\{v_1, ..., v_n\}$ represents its vertices and $E=\{e(v_i,v_j)\}$ the connections between vertex pairs. Here we focus on undirected and weighted networks, which means that $e(v_i,v_j) = e(v_j,v_i) \in \mathbb{R}$. Consider an $n$-by-$n$ matrix $W$ as a representation of the set of edges $e$, meaning that $W_{i,j} = e(v_i,v_j)$ (a.k.a.~adjacency matrix). One possible metric that can be computed from $W$ is the degree of a node $i$: 
%In the following, we describe in depth the fundamentals and properties of Complex Networks according to Network Science.
\begin{equation}\label{eq:strength}
    s(i) = \sum_j W_{i,j}\,,
\end{equation}
which, in our case, is referred to as weighted degree, or simply strength (since it is a weighted network). This metric, although simple, estimates node centrality and can be used to tackle important structural properties of networks. 
%Some known topological patterns usually found in real-world networks are the small-world \cite{watts1998collective} and scale-free \cite{barabasi1999emergence} properties, which describe the existence of hubs (nodes with many connections) and high information flow throughout the network.
It has been shown that several real-world networks exhibit a power-law distribution $P(s) = s^{-\gamma}$. This property describes scale-free networks~\cite{barabasi1999emergence} and has been found in many real-world networks of key scientific interest, from protein interactions to social networks and from the network of interlinked documents that make up the WWW to the interconnected hardware behind the Internet~\cite{barabasi2009scale}. Several other types of centrality measures can be computed from a network, and we refer the reader to~\cite{costa2007characterization}.

\subsection{%Related Work - 
Network Science approaches to Artificial Neural Networks}

Some works have already combined concepts from Network Science and ANNs, both for shallow and deep models. Shallow deep-belief networks were analyzed using Network Science tools in~\cite{testolin2020deep}, where correlations were found between the structure and function of neurons. They considered the correlation between the receptive field of neurons and its structural properties computed using centrality measures. Similarly, \cite{zambra2020emergence} analyzes the emergence of network motifs (a group of neurons with a particular connection pattern) during training of small fully-connected ANNs applied to synthetic data. They argue that the network topology can be strongly biased by choosing an appropriate weight initialization. Similar findings~\cite{scabini2021structure,la2021characterizing} also suggest that the structure and performance of ANNs are related. In~\cite{scabini2021structure}, a wide range of random networks is analyzed on the basis of several centrality measures computed for hidden neurons, where six neuronal types emerge during training in different tasks. From a statistical point of view, \cite{scabini2021structure} was the first work to explore a wide range of models to account for the variance caused by random weight initialization.

Although the previously discussed works provide interesting insights into the structure and dynamics of ANNs, they lack direct connections to efficient practical tools which could be integrated with state-of-the-art ANN construction, training, optimization, etc. On the other hand, some works have attempted this in specific scenarios. In~\cite{erkaymaz2016impact} the authors proposed a small-world ANN that has a better performance than a traditional structure for the task of diagnosing diabetes. The concept of weight sparsity was explored in~\cite{gray2017gpu} to obtain small-world properties on ANNs, allowing to train wider networks without incurring a quadratic increase in the number of parameters. Another work~\cite{xie2019exploring} explores random ANN architecture generation based on Network Science models by considering layers as nodes, and show that they can achieve competitive performance compared to optimized hand-crafted architectures. Similarly, \cite{you2020graph} proposes to analyze the graph structure of ANNs as a way to both generate new random architectures for computer vision and to correlate its complex network properties with performance. They suggest that performance is approximately a smooth function of the clustering coefficient and average path length (considering a relational graph modeling). A common trend among the mentioned works is that they do not consider a reliable sample size to account for the variance caused by the degrees of freedom during initialization and training.

%Going in a different direction, \cite{florindo2021VisGraphNet} consider the modeling of the output of vision architectures using complex networks, and achieve improved performance for texture analysis. In this particular case, the methodology needs trained models and works as some sort of transfer-learning to specific domains. Another approach worth mentioning is the use of Network Science to model images and, consequently, using randomized ANNs to learn their structural properties applied for texture analysis \cite{ribas2020fusion}. Here the idea goes differently: ANNs are applied over data represented as graphs (or complex networks), which is similar to the functioning of Graph Neural Networks \cite{wu2020comprehensive}. 

\section{%Network Science for 
Improving ANN Random Initialization}\label{sec:method}
\subsection{Weighted Node Degree}
We propose a practical method based on Network Science for ANN random weight analysis and improvement. The method is based only on the weight matrices of the model; therefore, we represent ANNs as collections of real matrices $\theta=\left( W_{n_l, n_{l+1}}\right)_l$. Each matrix $W$ is handled individually, and represents an undirected, weighted, and bipartite network
\begin{equation}
    G(W_{n_l, n_{l+1}}) = (V_l\times V_{l+1}, E)\,,
\end{equation}
where $V_l$ and $V_{l+1}$ are the sets of nodes representing the input and output neurons, respectively. The set of edges can be simply represented as the weight matrix
\begin{equation}
    E=W_{n_l, n_{l+1}}\,,
\end{equation}
where $E(i,x)$ denotes the connection weight between neuron $i \in V_{l}$ and $x \in V_{l+1}$. 
This notation naturally covers fully connected (FC) layers, and can also be extended to convolutional layers $W_{w,h,z,o}$ by expressing them as 2-dimensional matrices $W_{n_l, n_{l+1}}$ with shape $n_l = whz$ by $n_{l+1} = o$, where $w$ and $h$ are the spatial dimensions (width and height) of a filter, $z$ the number of input channels, and $o$ the number of output channels. In the graph analogy, this means that the filter dimensions represent the input neurons, and the number of filters the output neurons. This can also be extended to convolutional layers with 3 or more dimensions by flattening the additional (trailing) dimensions accordingly. In other words, we can express weight matrices in two dimensions for most types of ANN layers, meaning that our method can be applied in any such case. %In fact, most weight initializers from the literature approach weight matrices in this fashion. %Our analysis based on Network Science then considers ANN computational graphs as a set of 2-dimensional matrices or a set of input-output neuron pairs as complete bipartite graphs.

Given $G(W)$ for a given weight matrix, one possible approach to analyze its structure and dynamics is through node centrality. Since the network is a complete bipartite graph, this means that most of the centrality measures cannot provide meaningful information, and also that cost increases significantly if it depends on the number of edges. In this sense, and also following findings of previous works~\cite{scabini2021structure,la2021characterizing}, we resort to one of the simplest centrality measures: the (weighted) node degree, which we refer to as node strength ($s(i)$ in Eq.~\ref{eq:strength}). Results indicate that the strength distribution of small ANNs after training allows for distinguishing between models with different test performances and this on different vision tasks~\cite{scabini2021structure}. These findings suggest that the strength of trained neurons is related to the performance of ANNs. This measure can be considered as a direct estimate of the average firing rate of artificial neurons $f_i(\sum_j W_{i,j} X_j$) if we consider a linear activation function, no bias and i.i.d.~inputs.

\subsection{The strength of randomly initialized networks}
We consider randomly initialized networks, meaning that for each layer the corresponding matrix $W$ is generated by a given random weight initializer.
% B: this looks a bit strange to me, what is Omega, real space?, W belongs to R^{....}; L: Omega is supposed to be a random sampling method, I saw this notation in a paper but there is probably a better way to define it 
The strength of an artificial neuron $s$ in such network $G(W)$ is obtained by summing the random weights of the edges connected to it. The central limit theorem guarantees that the sum of independent random variables is normally distributed, even if the original sample space is not normally distributed. This means that the strength distribution $P(s)$ of layers in randomly initialized networks will approximately follow a normal distribution. This characteristic is to be expected for random graphs~\cite{erdos1959randomgraphs}. We can also estimate the variance of $s$ for a layer $l$ through the law of total variance by
\begin{equation}
  \sigma^2(s) = \sigma^2(W) n_l\,,
\end{equation}
\emph{i.e}, the variance of the sum of independent random weights (strength) is the sum of the variances of the weights. 
Since all weights have the same variance, given by a weight initializer, we can simply multiply this variance by the number of neurons. 

\begin{figure}[!htb]
    \centering
    \subfigure[Weight distributions generated by initialization methods (for sizes $n_{l} = n_{l+1} = 1024$).]{\includegraphics[width=0.33 \linewidth]{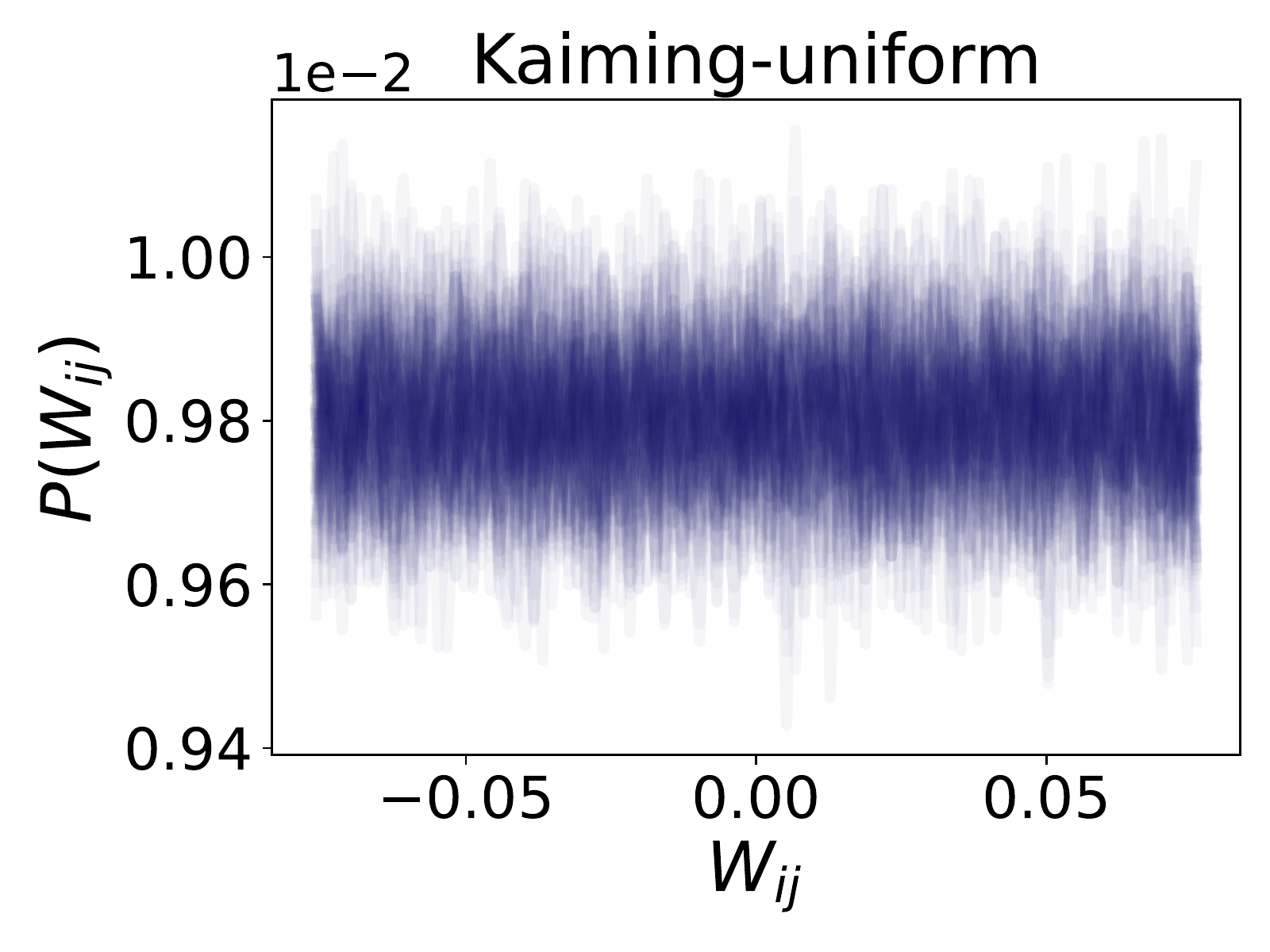} \includegraphics[width=0.33 \linewidth]{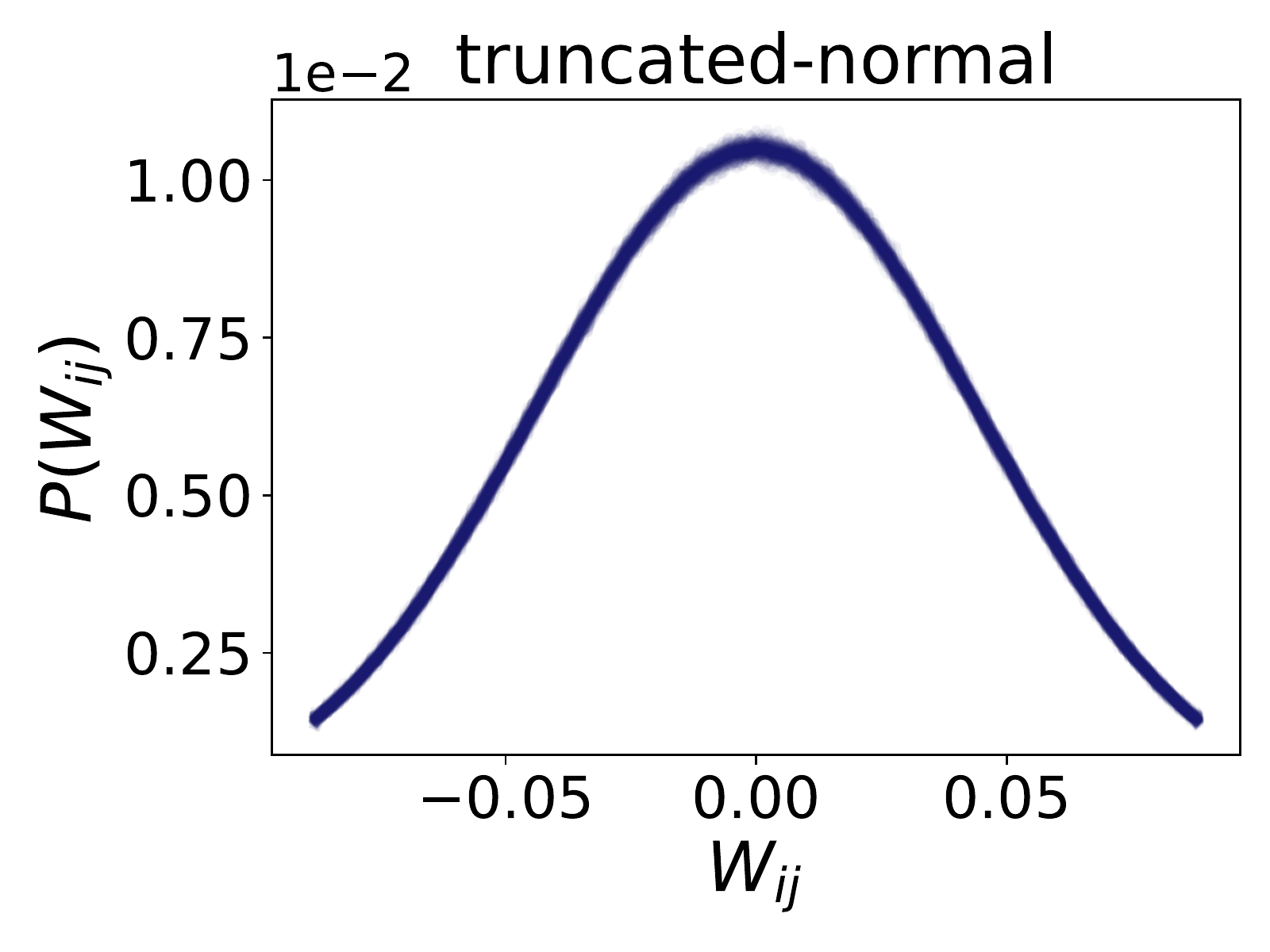} %\includegraphics[width=0.32 \linewidth]{orthogonal_weights.pdf}
    } \\
    
    \subfigure[Strength distributions for the above cases (a), including the result after applying the proposed PA rewiring.]{\includegraphics[width=0.33 \linewidth]{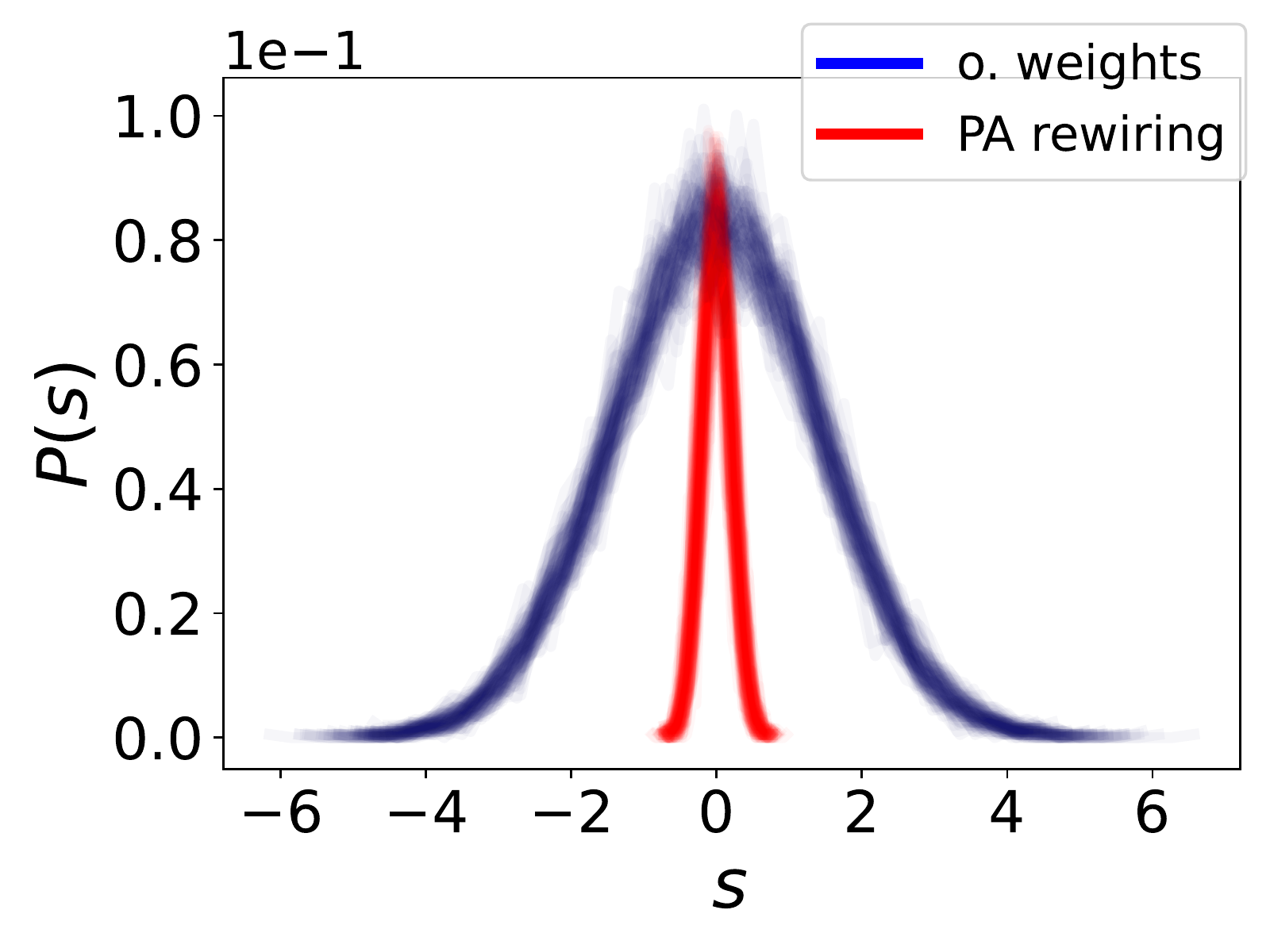} \includegraphics[width=0.33 \linewidth]{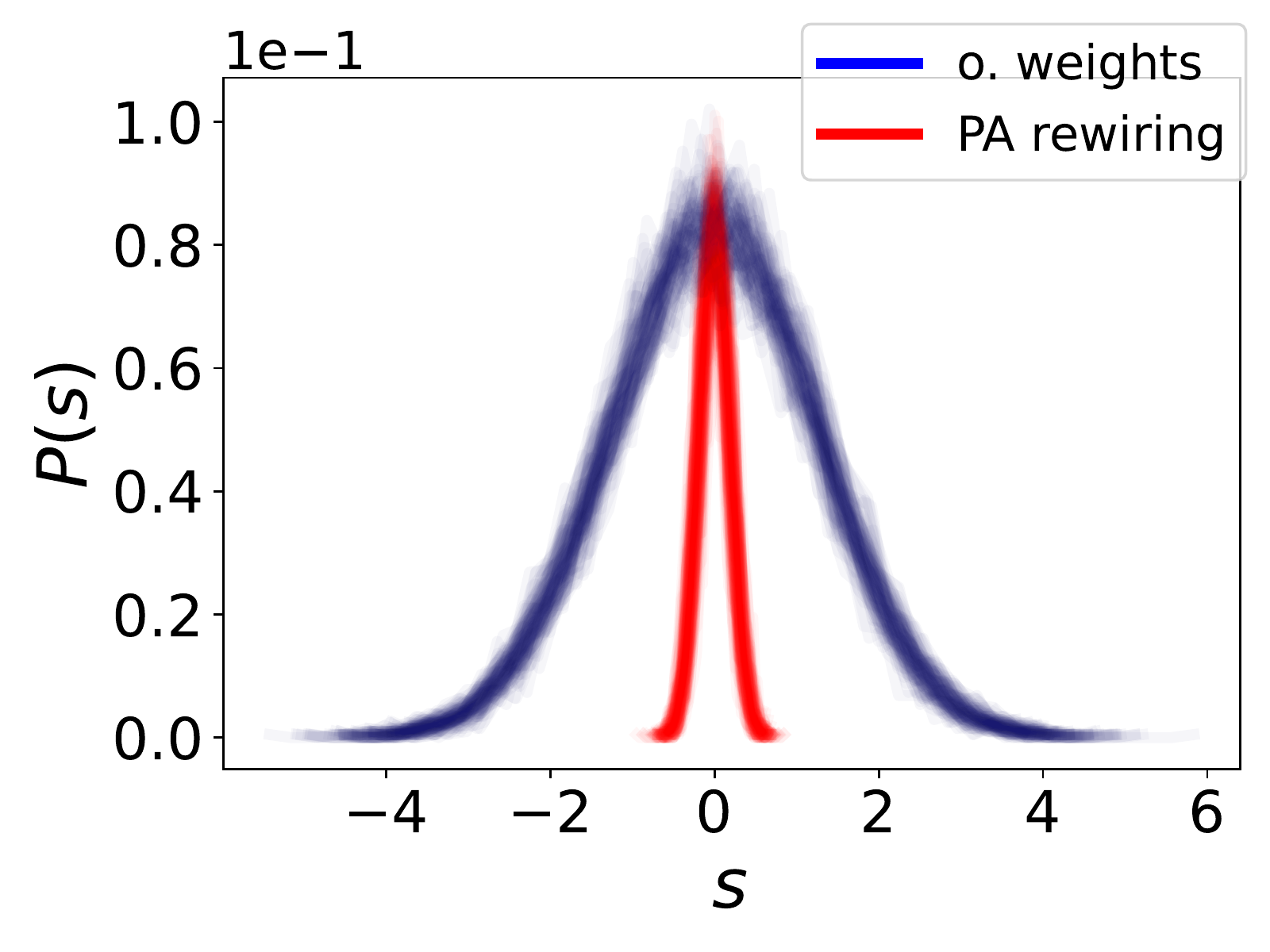} %\includegraphics[width=0.32 \linewidth]{orthogonal_strength.pdf}
    } \\
    
    \subfigure[Maximum strength distributions for different layer sizes.]{\includegraphics[width=0.33 \linewidth]{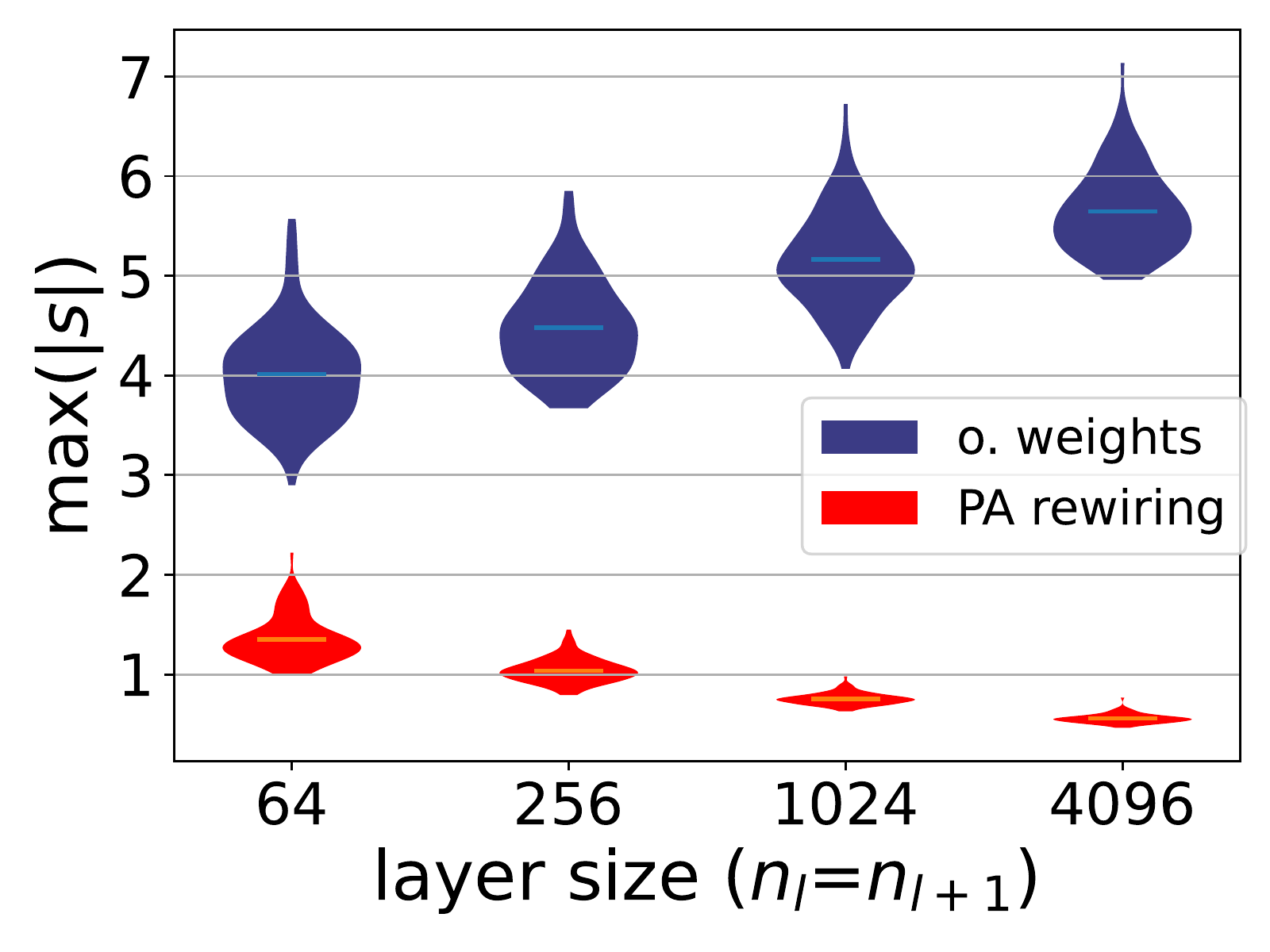} \includegraphics[width=0.33 \linewidth]{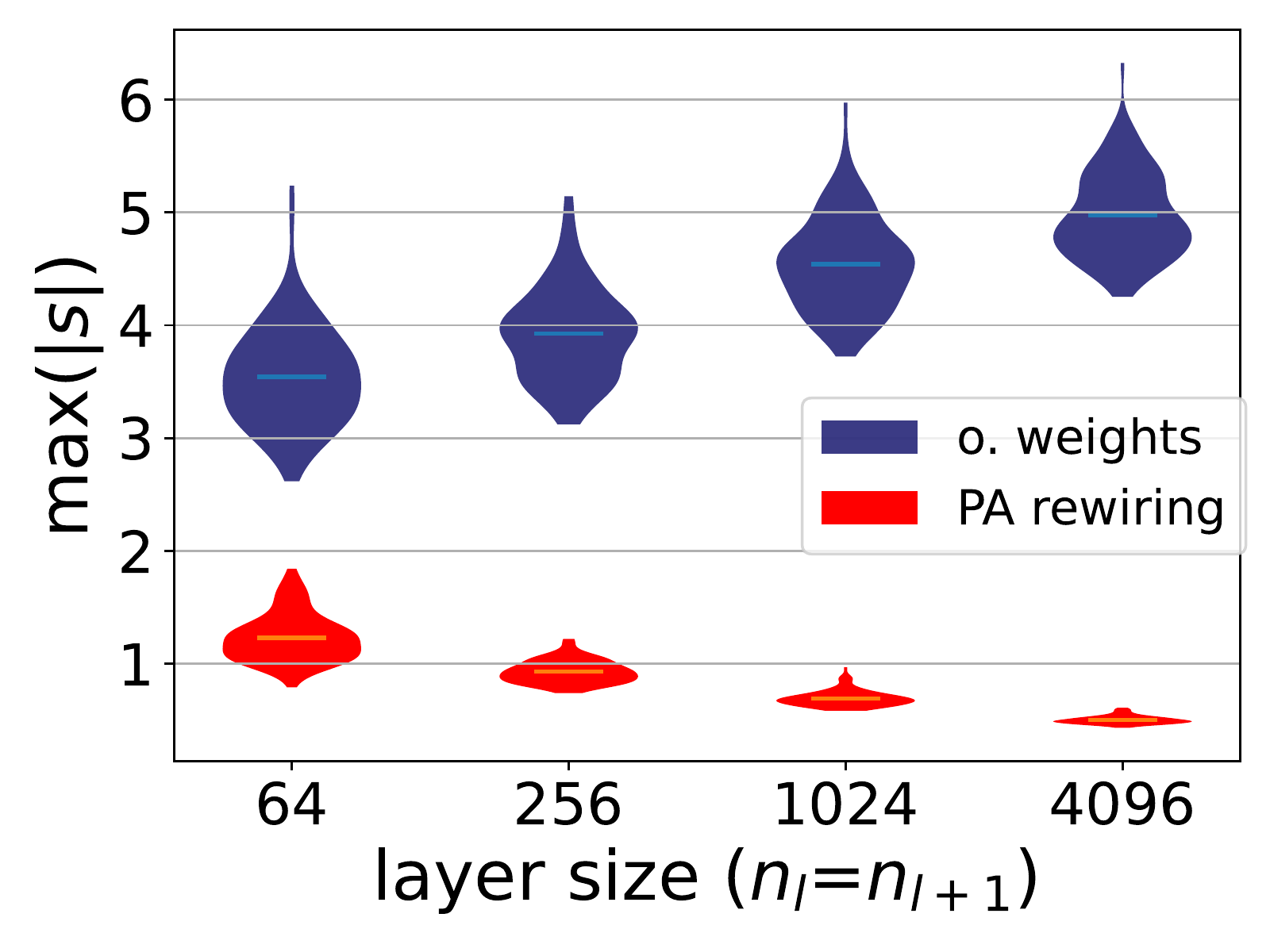} %\includegraphics[width=0.32 \linewidth]{orthogonal_maximum_degree.pdf}
    } \\
   
    \caption{\label{fig:degree_dispersion}Properties of random weights generated with literature methods (original weights, blue) or modified by the proposed PA rewiring method (red). We show the generated weights (100 trials) (a), their strength distributions before and after PA rewiring (b), and the distribution of maximum absolute strength observed in each case for different layer sizes (c).}
\end{figure}

Although we can estimate $\sigma^2(s)$, the bounds for the tails of the distribution $P(s)$ are indefinite (tend to $\pm \infty$), independently from the weight distribution shape (uniform, normal, truncated, etc). This means that even if the weights are sampled from a truncated distribution, outlier neurons will emerge in the strength distribution. To illustrate this, we generate several random layers with the Kaiming method~\cite{he2015delving} using both the uniform and a truncated version of the normal distribution (truncating at $\pm 3\sigma$). Results for the different methods and layer sizes are shown in Fig.~\ref{fig:degree_dispersion}. We first show 100 weight distributions for a FC layer of sizes $n_l = n_{l+1} = 1024$ and its strength distribution $P(s)$ (a-b, blue lines). One can notice the resulting long-tailed strength distribution for literature initializers. Figure~\ref{fig:degree_dispersion} (c, blue) also shows that the maximum absolute strength produced by literature methods increases with layer size, even though these methods consider this size to define the sampling interval. Moreover, the maximum distribution (c, blue) is also long-tailed, asymmetric, and has a considerable high variance. This means that neurons with highly negative and positive strengths emerge during random initialization in a complex way. Intuitively, these effects are also expected to happen with most other initializers (e.g., the original Kaiming-normal method).

The results shown above are not entirely unexpected, since they are well known in probability theory as the result of summing an increasingly higher number of random variables. Moreover, outlier nodes with extreme connectivity are a common property of real-world networks with the scale-free characteristic~\cite{barabasi1999emergence}, \emph{i.e.}, the presence of hubs causing the strength distribution to follow a power-law. Such behavior is expected for ANN neurons after training~\cite{scabini2021structure,la2021characterizing}. However, there is no evidence that such structure is beneficial at random initialization. We argue that hub neurons at initialization may harm the training performance and gradient flow, since they may dominate the layer signal both at forward and backward passes. This is intuitive to imagine if we consider a ReLU network, where a neuron with a highly positive strength may generate signals with a high magnitude, while a highly-negative strength may cancel a significant part of the signal. Such cases may cause problems of neuronal saturation, or exploding/vanishing gradients. In this sense, one would expect the initial strength of neurons to be correctly scaled for signals to propagate uniformly, and for more neurons to be able to learn collectively. In the following, we discuss our proposed method based on these observations.

\subsection{Rewiring % Random Networks 
 through Preferential Attachment}

We hypothesize that the high neuronal strength observed with literature initializers may not be ideal for ANN initialization. To modify their strength distribution, we consider a well-known model from Network Science, the scale-free network~\cite{barabasi1999emergence}. The algorithm to generate scale-free networks is based on a strength preferential attachment (PA) rule, which consists of increasingly including new nodes and edges into the network giving higher priority to hubs ($P(i) = \frac{s(i)}{\sum_j s(j)}$) when wiring them. This greedy iterative process causes a small fraction of nodes to receive a large portion of the network connections, creating the so-called hubs, while the majority of nodes keep a small strength.

For ANN initialization, we propose to avoid the scale-free structure usually present in real-world networks. First, consider that randomly initialized layers may contain negative strengths and connections (the most common approach for ANN initialization). Given a weight matrix $W_{n_l, n_{l+1}}$, initialized by some random initializer, consider $i$ as indices for neurons in the input layer $l$ and $x$ as indices for neurons in the output layer $l+1$. Let us start by assuming that only one of the neurons $x$ in layer $l+1$ is connected to all other neurons in layer $l$, which we will refer to as the first neuron $x=1$. In this sense, we can compute the current temporary strength $s_{t=1}$ of a neuron $i$ in the first layer, at the initial iteration $t=1$, by $s_{t=1}(i) = W_{i,1}$, since each neuron $i$ will contain only a single connection coming from $x$. We then propose the following PA probability for these nodes to receive "new" connections
\begin{equation}
    P(i) = \frac{s_t(i) + |\min(s_t)| + 1}{\sum_j s_t(j) + |\min(s_t)| + 1}\,,
\end{equation}
which means that we shift the strength distribution to the positive side, and add 1 to avoid null probability. In this sense, the probability ranges from the lowest negative strength (smaller probability) to the highest positive strength (higher probability). Now we cycle through $t \in [2, n_{l+1}]$ iterations visiting a neuron $x$ at each iteration. The weights $W_{i,x}$, $i=1,\ldots, n_l$, in ascending order, are then rewired to nodes in layer $l$ according to the probabilities $P(i)$. In other words, neurons with negative strength have more chance to receive new positive connections, while the same goes for positive neurons and negative edges. At each iteration $t$, a new temporary strength is calculated after the last rewiring (previous iteration $t-1$), 
\begin{equation}
    s_t(i)=\sum_{x=1}^{t-1} W_{i,x} =  s_{t-1}(i) + W_{i,t-1}\,,
\end{equation}
along with the new probabilities $P(i)$. A new rewiring w.r.t.~the input neurons can then be performed with the weights of a new neuron $x=t$. After rewiring all neurons in layer $l+1$, this procedure will reorganize ANN weights so that the neuronal strength of input neurons is close to zero, as we show in Fig.~\ref{fig:degree_dispersion} (b-c, red). The whole process is also described in Algorithm~\ref{alg:method}. Note that we assume that $W$ is already filled with random values, which is needed for computing the strengths, and no weights are created in practice as we simply assume that existing weights are reorganized.

\begin{algorithm}\label{alg:method}
\caption{A python-like pseudo-code of the proposed PA rewiring method.}

% \begin{mintedbox}{python}
% import numpy as np
% def PA_rewiring(weights):
%     dimensions = weights.shape    
%     st = np.zeros(dimensions[1])
%     for neuron in range(1,dimensions[0]): 
%         st = st + weights[neuron-1]        
%         P = st + np.abs(np.min(st)) + 1
%         P = P / np.sum(P)   
%         targets = np.random.choice(a=[i for i in range(dimensions[1])], replace=False, 
%                                       size=dimensions[1],p=P) 
%         edges_to_rewire = np.argsort(weights[neuron]) 
%         weights[neuron, targets] = weights[neuron, edges_to_rewire]                
%     return weights \end{mintedbox}

   \SetKwInOut{Input}{Input}
    \SetKwInOut{Output}{Output}

    \underline{function PA$\_$rewiring} $(W)$\;
    \Input{An $n_l\times n_{l+1}$ weight matrix $W$ filled by some weight initializer}
    \Output{$W$ after rewiring}
    
    $s_1$ = zeros($n_{l}$)
    
    \For{$t$ from $2$ to $n_{l+1}$,}
      {
        $s_t = s_{t-1} + W[:, t-1]$
        % not nice that t is the loop variable, so that does s_t then mean on line 2; L: fixed, line 2 is t=1, and line 4 sums the previous s_t
        % B: should be W at t, not?
        %L: Its W at t-1 plus the previous strength (s_t-1)
      
        $P = s_t + \mathrm{abs}(\min(s_t)) + 1$
        
        $P = P / \mathrm{sum}(P)$
        
        targets = random$\_$choice([$i \in [1,...,n_{l}]$], prob=$P$, size = $n_{l}$)
        
        new$\_$edges = argsort($W[:, t]$)
        
        $W[\mathrm{targets}, t] = W[\mathrm{new}\_\mathrm{edges}, t]$
        
      }
      {
        return $W$
      }
\end{algorithm}

The PA rewiring process described performs the following action: for a layer and its corresponding weight matrix,
% B: here is a confusion, layer pair means 2 weight matrices according to your answer before
% L: I'll refer to it as layer = represented by a 2d weight matrix, with n_l inputs and n_{l+1} neurons
rewire its connections so that the strength distribution of the input neurons tends to zero. This procedure does not affect the strength of the neurons in the output layer, since we are rewiring just the end-point of the connection pointing to the input layer, for each output neuron individually. To achieve the same for the output neurons, the rewiring process can be repeated by simply reversing the targets: rewiring connections from each input neuron to the output neurons, making the strength of the output neurons also tend to zero. 

\begin{figure*}[!htb]
    \centering
     \includegraphics[width=0.8 \linewidth]{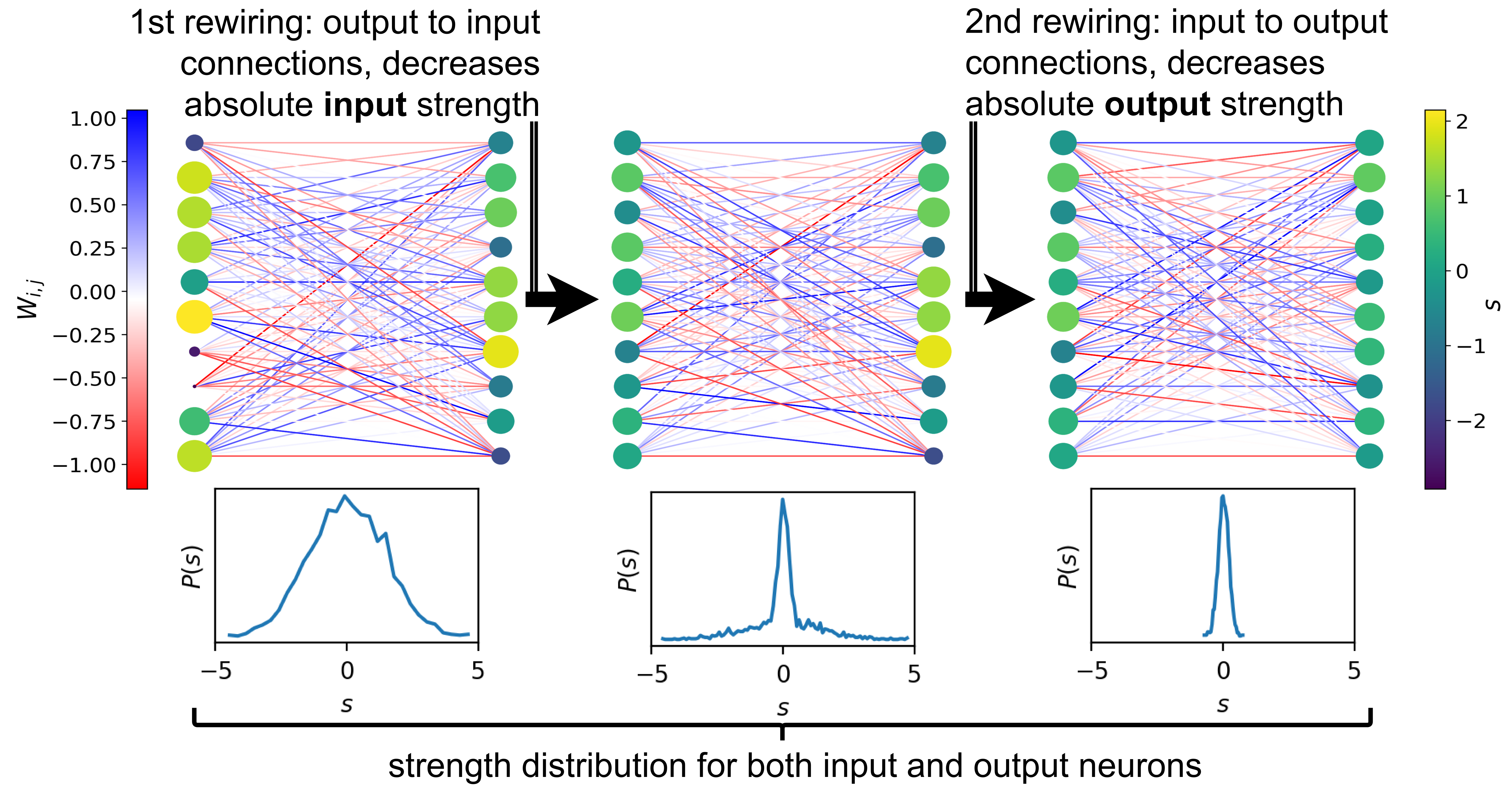}

    \caption{\label{fig:diagram}Illustration of the functioning of PA rewiring at reorganizing weights to decrease absolute neuronal strength. The size of the neurons represents their strength, from the most negative (smaller) to the most positive. We show a small layer with 10 neurons and 10 inputs, while the strength distributions are calculated for a layer with 1024 neurons and 1024 inputs.}
    % sounds strange: two layers of size 1024?; L: it means the number of input and output neurons (since each layer is represented by a 2D weight matrix)
    % maybe we have to describe the latter more clearly
\end{figure*}

In fact, the distributions shown in Fig.~\ref{fig:degree_dispersion} (b-c) after rewiring already consider this two-step procedure for layers with sizes $n_l = n_{l+1}$. The whole procedure is illustrated in Fig.~\ref{fig:diagram}, where one can observe how an initial long-tailed strength distribution, generated by a traditional initializer, vanishes when connections are reorganized with PA rewiring. Note that the two rewiring rounds are needed to effectively reduce both input and output strengths. In this sense, we are considering both signals to be equally important for ANN functioning. However, it is important to stress that we did not analyze the individual impacts of the input and output strengths, for instance by testing the rewiring individually for each of them. 
%Moreover, our strength measure is restricted to a single layer since one graph is modeled for each weight matrix, instead of modeling %sets of layers or the whole network. 
% I cannot make sense of the latter sentence; L: I mean that we are modeling each layer individually, not the whole network as a single graph. I included that in the sentence
% approach individual graphs (?) to (?) each weight matrix ??
% we'll discuss, I think you are stuck, trying to make it sound more complicated than it is
%In this sense, we also do not account for the second-order effects of the strength between layers or throughout the architecture. %However, these properties should be considered for future works and may help to derive more specific and effective methods.

\subsection{Computational complexity}

The cost of the PA rewiring algorithm mainly depends on the size of the layer one needs to rewire, \emph{i.e.}, the numbers of input and output neurons $n_{l}$ and $n_{l+1}$. The main loop iterates over all except one of the output neurons $x$ in layer $l+1$, in a total of $n_{l+1} - 1$ iterations.
Let us then estimate the cost of the operations inside the loop. The computation of the temporary strength $s_t$ (line 4) adds an array of $n_{l}$ connections to the previous stored array of strengths, \emph{i.e.}, a cost of $O(n_{l})$. To compute the PA probabilities, we need to find the minimum strength ($O(n_{l+1})$), add it to $s_t$ ($O(n_{l})$), and divide it by the sum of $s_t$ ($O(n_{l})$). Therefore, the computation of $P(i)$ entails a total cost of $O(3n_{l})$. The cost to select the target nodes (line 7) also depends on the number of neurons in the layer ($O(n_{l})$). Finally, the most expensive operation inside the loop is the sorting of the connections between $x$ and all other neurons in layer $l$, yielding $O(n_{l}log(n_{l}))$. There is an additional cost of $O(n_{l})$ to effectively rewire the connections, \emph{i.e.}, to swap the weight matrix positions (line 9). The total cost for the operations inside the loop is then $O(6n_{l} + n_{l}log(n_{l}))$. By nesting these operations with the outer loop and removing the constant, the total cost is
\begin{equation}
    O(n_{l+1}(n_{l} + n_{l}\log(n_{l})))
\end{equation}
or, if we consider that $n_{l} \approx n_{l+1} = n$ (average number of neurons per layer),
\begin{equation}
    O(n^2\log(n))\,.
\end{equation}

It is important to note that since our graph modeling approach depends on the number of weights $|W|$, $n$ will be considerably smaller for convolutional layers compared to dense (FC) layers.

\section{Experiments and Results}\label{sec:results}
In this section, we give a detailed description of the experimental setup and 
the results obtained. We focus on a more robust statistical evaluation rather than performing a single trial with several models on huge datasets, which is the most common approach in the literature. Instead, several repetitions are performed by varying random seeds for weight initialization. A total of 100 repetitions were performed for each case (combination of model, initializer, and task) with the first learning schedule, and 10 repetitions for the experiments with deeper models and data augmentation (Section~\ref{sec:deeper}). We believe that this is a more comprehensive and robust approach to assessing the individual impacts of random weights, which is our main goal (rather than seeking the highest possible performance). %Moreover, some experimental constraints were carried out due to limited computing resources.

\subsection{Experimental Setup}

The results presented in this section correspond to two different scenarios: a baseline case with a simple learning schedule that is less computationally expensive, and a more complex scenario (see Section~\ref{sec:deeper}). For the former case, we do not employ any data augmentation, which allows us to allocate all training data into GPU memory for faster computation. We consider SGD with momentum equal to $0.9$, batch size 128, and a cosine schedule~\cite{loshchilov2016sgdr} with an initial learning rate of 0.01 (annealed to
0, no restarting). We found that this configuration was the better overall choice to achieve convergence with the architectures and datasets analyzed in this first scenario. We test different models, initialization methods, and datasets, and also perform a statistical analysis (100 trials) of both model performance and training dynamics.
% B: case might be confusing here, not clear what it refers to
% L: I removed it, since we already described the cases before

% \begin{itemize}
%     \item MNIST \cite{lecun1998gradient}: This is one of the most known computer vision benchmarks, with grayscale 28x28 images representing handwritten digits from 0 to 9. It comprises 60k images for training and 10k for testing where the goal is to classify each digit.
    
%     \item FMNIST \cite{xiao2017fashion}: Fashion MNIST contains more detailed objects in comparison to MNIST and is composed of 28x28 grayscale clothing images, with the following classes: t-shirt, trouser, pullover, dress, coat, sandals, shirt, sneaker, bag, and ankle boots. The number of samples is the same as in MNIST.

%     \item CIFAR10 \cite{krizhevsky2009learning}: Comprises 32x32 colored images of 10 classes: airplane, automobile, bird, cat, deer, dog, frog, horse, ship, and truck. It is a more challenging classification problem as images are uncontrolled and object conditions and positions vary greatly. Contains a fixed split of 50k images for training and 10k images for testing.

% \end{itemize}

\textbf{Datasets:} To cover different tasks, we consider MNIST~\cite{lecun1998gradient}, FMNIST~\cite{xiao2017fashion}, and CIFAR10~\cite{krizhevsky2009learning}. Each dataset contains a fixed train and test set, and we use an additional split: a new validation set randomly sampled from the training set, with the same size as the test set, used for model selection. The test set is held out for evaluation at the end of the training. It is well known that the use of an additional validation data split
% actually, this is not a third split, but the second; no need to number
is highly important to assess the real capabilities of a model and to avoid a biased and overfitted selection.
However, this is not usually done in deep learning research since the ultimate goal is to maximize performance on a fixed test set, which is the case of the most common datasets (e.g.~the ones used here). In this scenario, there is a danger of overfitting to excessively re-used test sets~\cite{recht2019imagenet}. Therefore, the use of different validation sets has been discussed in the deep learning literature as a good practice for a more robust experimental setup~\cite{recht2019imagenet,picard2021torch,wightman2021resnet}. %Considering our present work, it is important to stress that our goal is not to achieve the highest possible performance but to strictly measure the effects of random weights with meaningful sample size. For instance, the use of an additional validation set consequently reduces the number of training samples, leading to reduced performance. Moreover, we apply the models on the real test set just once:  after finishing training, using the best performing model on the validation set.

\textbf{Weight initializers:} for the base weights we considered the following methods: both uniform and normal versions of the Kaiming method~\cite{he2015delving}, a truncated version of the Kaiming-normal method (truncating at $\pm 3\sigma$), and orthogonal weights~\cite{saxe2013exact,hu2020provable}. It is important to stress that these techniques were selected due to their similarity to the proposed method, \emph{i.e.}, they work locally in a given layer without external knowledge from the data or the architecture (except for the activation function needed to define the variance of the weights). For each initializer, 100 trials are performed, \emph{i.e.}, 100 random seeds are used to initialize each model, in each experiment. The proposed PA rewiring method is applied on the original weights generated by the same seeds.
% unclear: operations on weights are applied on weights?
% L: I mean that the rewiring is applied over the same random weights generated by the literature methods (achieved using the same seed) 
To quantify the individual effects of different random weights, all other stochastic factors involved in the model construction and training are fixed using a global seed. Data is randomly split into training and validation once, and the same splits are maintained throughout all trials. The training random shuffle, \emph{i.e.}, the order in which the training samples are presented to the model, and the random batch generation are also maintained for all trials.

\textbf{Statistical evaluation:} to obtain statistically meaningful results, we consider the average and standard deviation for all metrics, as well as the median and median absolute deviation. Moreover, when comparing the averages of two methods, we also perform a t-test considering the null hypothesis that both cases have identical averages. When comparing the medians of two methods, the Kruskal-Wallis H-test test is performed to check the null hypothesis that their medians are equal. In both cases, we consider a statistically significant difference when $p<0.05$ (more than $95\%$ confidence). Colored table cells represent whether or not the statistical test detected a significant improvement (green), a significant weakening (red), or if the null hypothesis was confirmed with $p\geq0.05$ (gray). Statistical tests are always performed w.r.t.~a literature initializer compared to some weight operation (such as the PA rewiring method). 
% but not mentioned that improvement/weakening refer to the rewiring method.
% L: yes, but there are some cases where I do statistical tests with other methods too. I included an additional sentence  

\subsection{The effects caused by the neuronal strength of random networks}

\begin{figure}[!htb]
    \centering
    \includegraphics[width=0.43 \linewidth]{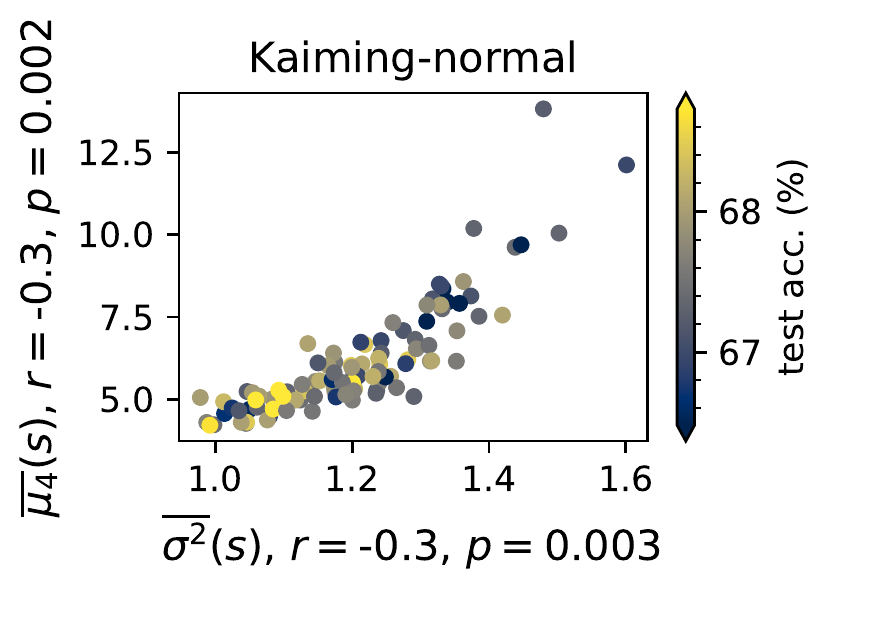}
    \includegraphics[width=0.43 \linewidth]{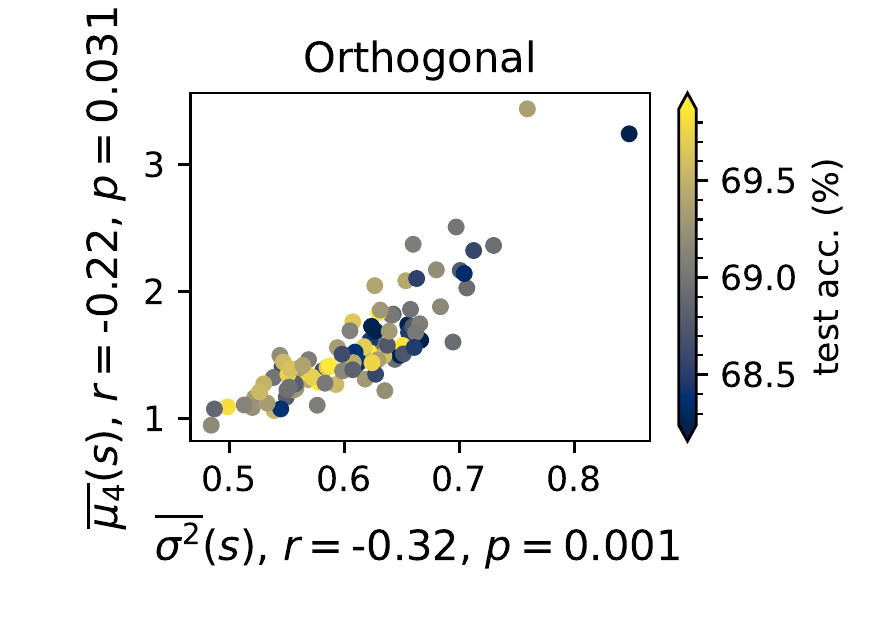}
 
    \caption{\label{fig:statistics}The statistical properties of the neuronal strength distribution (average variance and fourth central moment) of a set of 100 LeNet models generated with each random weight initializer, trained on CIFAR10. Statistics are computed before training, and colors represent the final test accuracy after training; $r$ and $p$ are the Pearson correlation coefficient and corresponding $p$-value.}
\end{figure}

To highlight the impact of neuronal strength on ANN performance, we start with a baseline experiment as a sanity check. A LeNet architecture~\cite{lecun1989backpropagation} is considered, with four layers (two convolutional and two FC), ReLU neurons and additional layer sizes of 20 and 50 (convolutional filters), and 800 and 500 (FC layers), respectively. This configuration was employed considering that it has the same number of convolutional and FC layers, and the layer widths were increased to compensate for the lack of one layer (common in the LeNet-5 version). A hundred random LeNet models are then generated, each with a different random initialization. Before training, we compute statistical properties from their strength distribution. The models are then trained on CIFAR10 with everything else fixed: the same training hyperparameters, data splits, etc. Therefore, the only degree of freedom is the random weight organization, while the shape of the distribution is the same since we are considering a single initializer in each experiment. Figure~\ref{fig:statistics} shows the spatial distribution of the networks considering the average variance $\overline{\sigma^2} (s)$ (between all layers) and average fourth central moment $\overline{\mu_4} (s)$ (non-normalized kurtosis) of their strength distributions. Colors represent the final test accuracy after training; $r$ and $p$ are the Pearson correlation coefficient and the corresponding $p$-value testing the null hypothesis, \emph{i.e.}, whether the correlation coefficient is in fact zero. It is possible to notice a statistically significant ($p<0.05$) negative correlation of $\approx -0.3$ between both measures before training and the model's performance at the end of training. This means that strength distributions with a lower variance and tails will likely provide higher performance after training. It is also possible to notice that the Orthogonal initializer produces strength distributions with smaller tails and lower variance, which seems to reflect its improved overall accuracy compared to the Kaiming-normal initializer. The fact that such measurable patterns can be detected in random weights corroborates our hypothesis that the strength distribution is an important factor when initializing ANNs.

\begin{figure}[!htb]
    \centering
    \subfigure[MNIST]{
      \includegraphics[width=0.4 \linewidth]{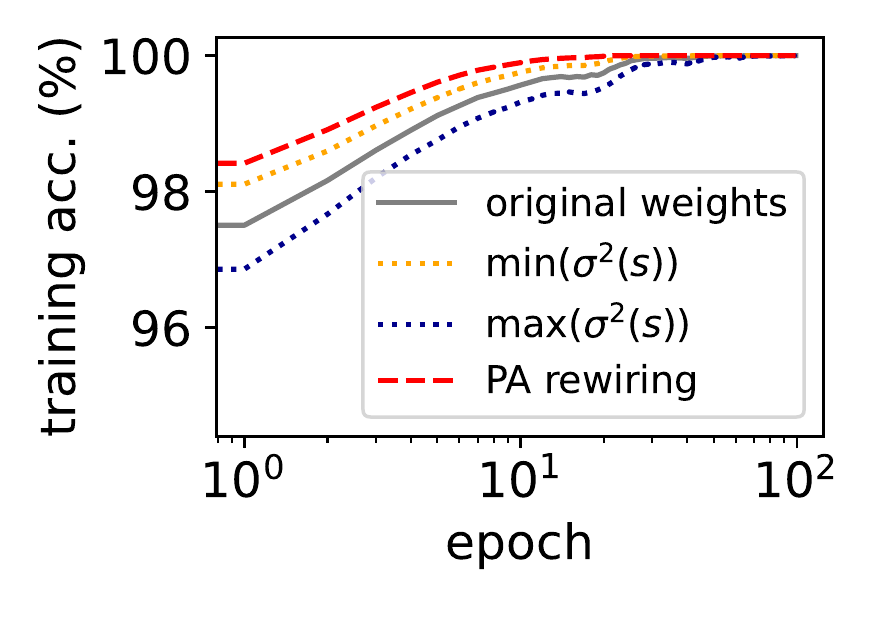}
     \includegraphics[width=0.4 \linewidth]{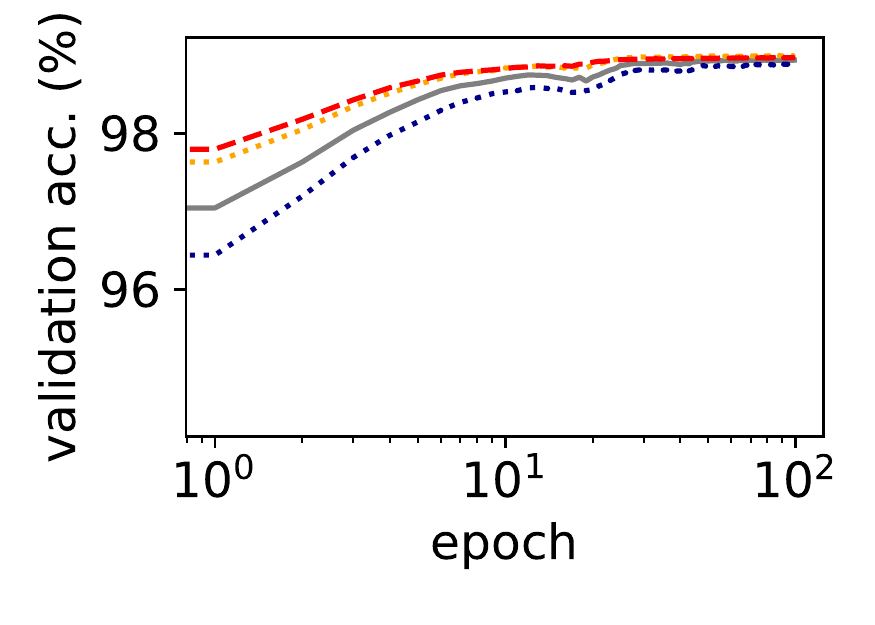}
     }
     
    \subfigure[CIFAR10]{
      \includegraphics[width=0.4 \linewidth]{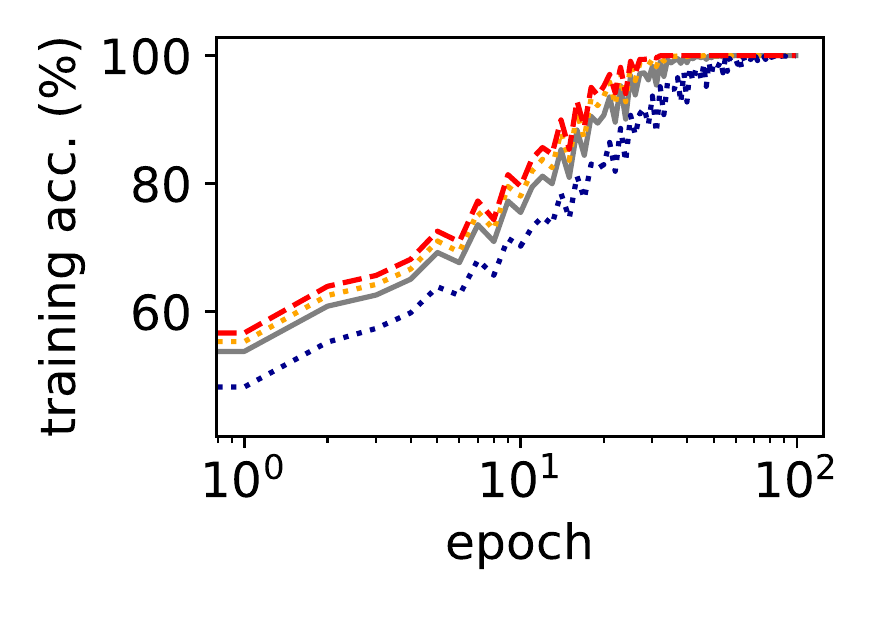} 
       
     \includegraphics[width=0.4 \linewidth]{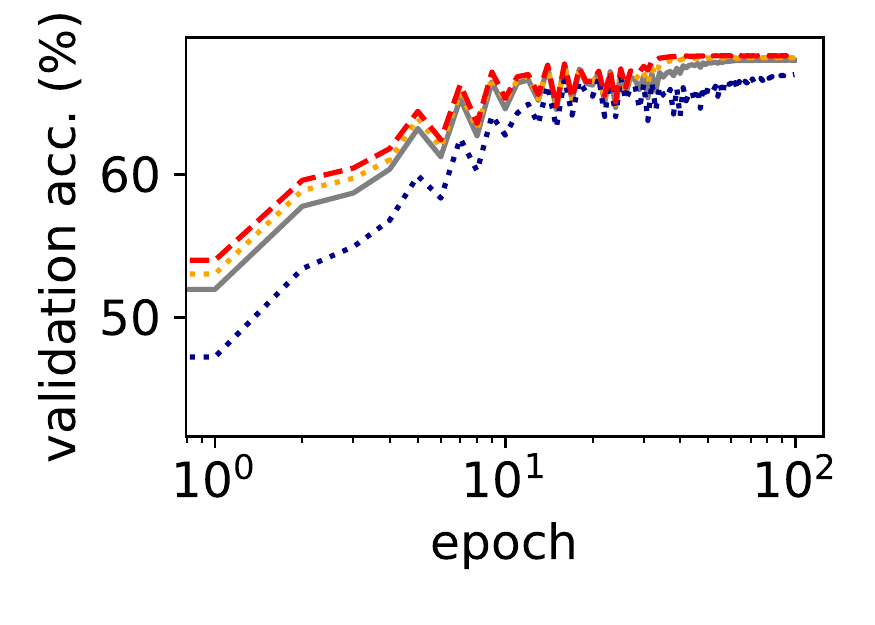}
     }

    \caption{\label{fig:dynamics1}The average training dynamics of 100 LeNet networks on MNIST and CIFAR10 with the Kaiming-normal initialization (grey line) compared to applying different operations on the original weights. The x-axis (epochs) is on a logarithmic scale for better visualization, since higher increases happen at the first training epochs.}
\end{figure}

We propose to explore the strength distribution of random ANNs as a way to improve initialization. First, a naive approach is considered as a baseline: given a network and a random weight initializer, for each layer, repeat the initialization process $K$ times and select one layer from this set that satisfies some desired property. The minimum and maximum strength variance ($\sigma^2(s)$) of the layer are considered as selection criteria. It is important to stress that we are not selecting the tails of the average distribution for the whole model, as shown in Figure~\ref{fig:statistics}. Instead, this is a simple random search applied to each layer of the model independently, producing layers that are extremely unlikely to appear simultaneously in a single model by chance. In addition, we also apply the proposed PA rewiring method to the original weights (same seeds). Figure~\ref{fig:dynamics1} shows the average training dynamics of LeNet networks by considering the described approaches on the MNIST and CIFAR10 datasets. It is possible to notice a symmetric behavior: while maximizing the strength variance decreases both training and validation accuracy during training, minimizing it increases both metrics, surpassing the original weights. Moreover, the proposed method provides a greater improvement than simply randomly reducing the strength variance. In other words, these procedures create models with a better initial parameter space, as the results at epoch 1 show, and this superiority seems to persist during training.

Table~\ref{tab:strength_impacts} shows statistical metrics of the results in Figure~\ref{fig:dynamics1}, and also of the final test accuracy of the models. The final model is selected considering the highest validation accuracy obtained during the 100 training epochs. A considerable statistical significance is achieved when confirming that a higher strength variance decreases model performance, while lower values increase it. It is also possible to observe that it similarly affects the convergence time. Another interesting effect is that the standard deviation of most metrics increases when maximizing the strength variance. On the other hand, when minimizing the strength variance or using the proposed PA rewiring method, the standard deviations decreases, corroborating a more stable scenario. PA rewiring achieves the best results overall, both regarding training performance and convergence time, and also for the performance after training. These results corroborate two hypotheses: a lower neuronal strength at random initialization is important for better training dynamics, as random minimization shows; and the organization of the weights is also important since we achieve both properties using PA rewiring and better results. 

\begin{table*}[!htb]
	\centering
	\caption{\label{tab:strength_impacts}Statistical metrics computed for 100 LeNet models trained on MNIST and CIFAR10, with different weight initializations. Colors represents statistical tests ($p<0.05$) performed w.r.t.\ the base weights (Kaiming-normal), with green meaning a positive difference, and red representing a negative difference.}
	
	\begin{tabular}{cc|ccc|cc}
		\hline
    % 	&&\multicolumn{3}{c|}{convergence (epochs)} &\multicolumn{3}{c}{test accuracy ($\%$)} \\
    % 	& method &min&average&max&min&average& max \\
		& & \multicolumn{3}{c|}{training metrics} & \multicolumn{2}{c}{test acc.}  \\
		
		dataset&weights&epoch 1 acc.&epoch 1 val. acc.&convergence&average&median\\ 
        \hline
\multirow{4}{*}{\rotatebox[origin=c]{90}{MNIST}}
& Kaiming-normal&95.93{\tiny$\pm$2.31}&95.54{\tiny$\pm$2.26}&16.10{\tiny$\pm$6.59}&99.01{\tiny$\pm$0.13}&99.03{\tiny$\pm$0.13} \\
& $\max(\sigma^2(s))$&\cellcolor{red!45}94.67{\tiny$\pm$2.68}&\cellcolor{red!45}94.35{\tiny$\pm$2.64}&\cellcolor{red!45}20.95{\tiny$\pm$8.42}&\cellcolor{red!45}98.95{\tiny$\pm$0.18}&\cellcolor{red!45}98.98{\tiny$\pm$0.11} \\
& $\min(\sigma^2(s))$&\cellcolor{green!45}96.93{\tiny$\pm$0.61}&\cellcolor{green!45}96.53{\tiny$\pm$0.62}&\cellcolor{green!45}13.06{\tiny$\pm$4.30}&\cellcolor{green!45}99.10{\tiny$\pm$0.08}&\cellcolor{green!45}99.10{\tiny$\pm$0.06} \\
& PA rewiring&\cellcolor{green!45}97.37{\tiny$\pm$0.31}&\cellcolor{green!45}96.85{\tiny$\pm$0.32}&\cellcolor{green!45}10.22{\tiny$\pm$2.40}&\cellcolor{green!45}99.10{\tiny$\pm$0.07}&\cellcolor{green!45}99.10{\tiny$\pm$0.05} \\
\hline

\multirow{4}{*}{\rotatebox[origin=c]{90}{CIFAR10}}
& Kaiming-normal&50.17{\tiny$\pm$1.60}&49.09{\tiny$\pm$1.34}&10.95{\tiny$\pm$1.98}&67.59{\tiny$\pm$0.65}&67.57{\tiny$\pm$0.65} \\
& $\max(\sigma^2(s))$&\cellcolor{red!45}43.37{\tiny$\pm$6.85}&\cellcolor{red!45}42.95{\tiny$\pm$6.66}&\cellcolor{red!45}15.75{\tiny$\pm$3.23}&\cellcolor{red!45}66.81{\tiny$\pm$0.74}&\cellcolor{red!45}66.76{\tiny$\pm$0.51} \\
& $\min(\sigma^2(s))$&\cellcolor{green!45}52.08{\tiny$\pm$0.93}&\cellcolor{green!45}50.76{\tiny$\pm$0.91}&\cellcolor{green!45}9.86{\tiny$\pm$1.02}&\cellcolor{green!45}67.77{\tiny$\pm$0.46}&\cellcolor{green!45}67.75{\tiny$\pm$0.34} \\
& PA rewiring&\cellcolor{green!45}53.30{\tiny$\pm$0.56}&\cellcolor{green!45}51.57{\tiny$\pm$0.58}&\cellcolor{green!45}9.60{\tiny$\pm$0.80}&\cellcolor{green!45}68.03{\tiny$\pm$0.53}&\cellcolor{green!45}68.07{\tiny$\pm$0.37} \\
\hline
	\end{tabular}
\end{table*}

\begin{figure*}[!htb]
    \centering
    \subfigure[MNIST]{
    \includegraphics[width=0.37 \linewidth]{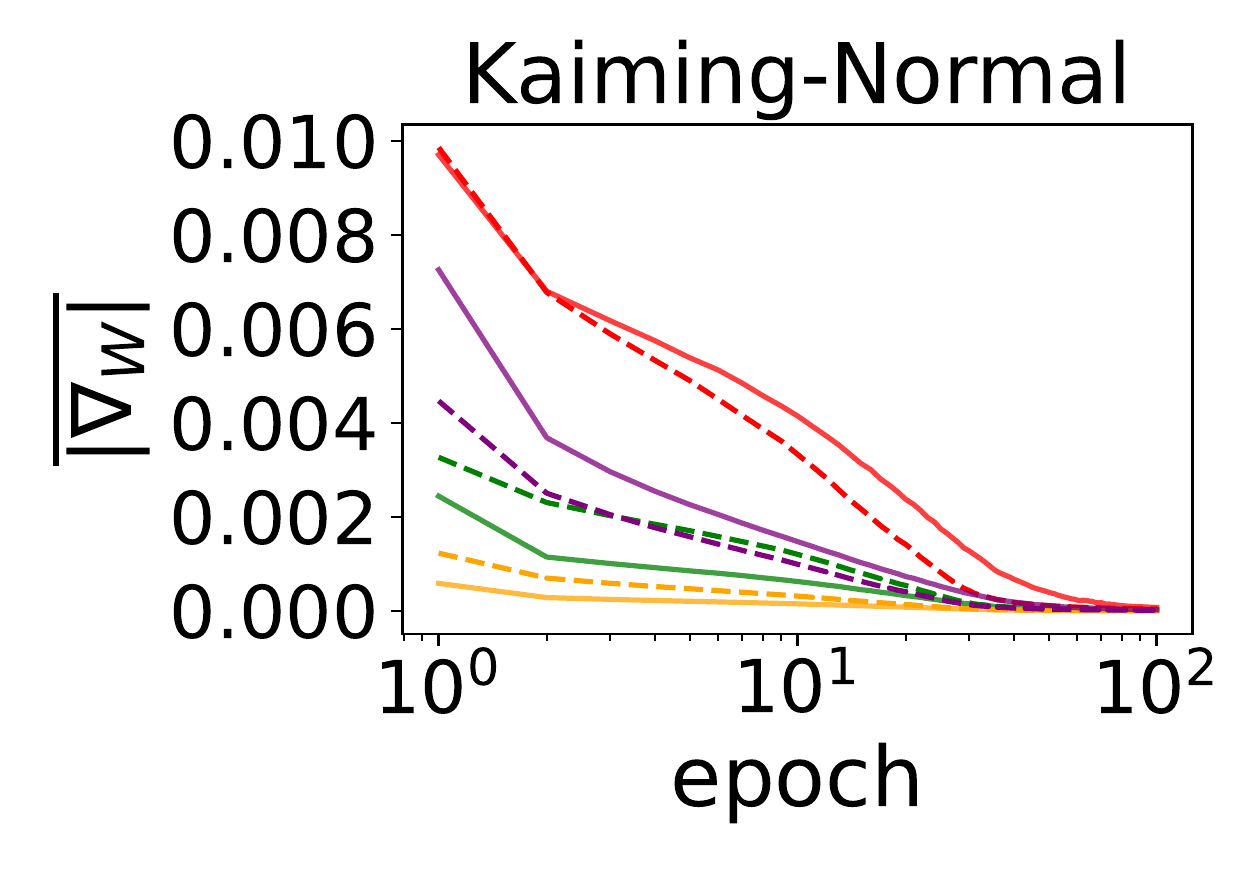} \includegraphics[width=0.2 \linewidth]{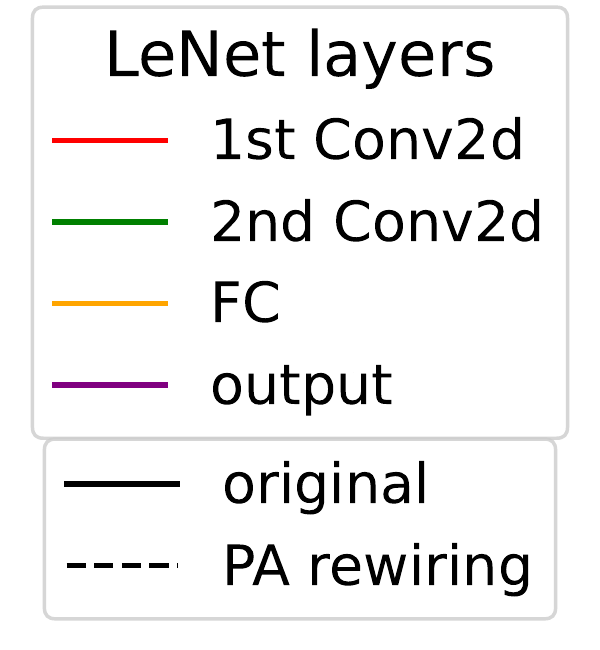}
    \includegraphics[width=0.37 \linewidth]{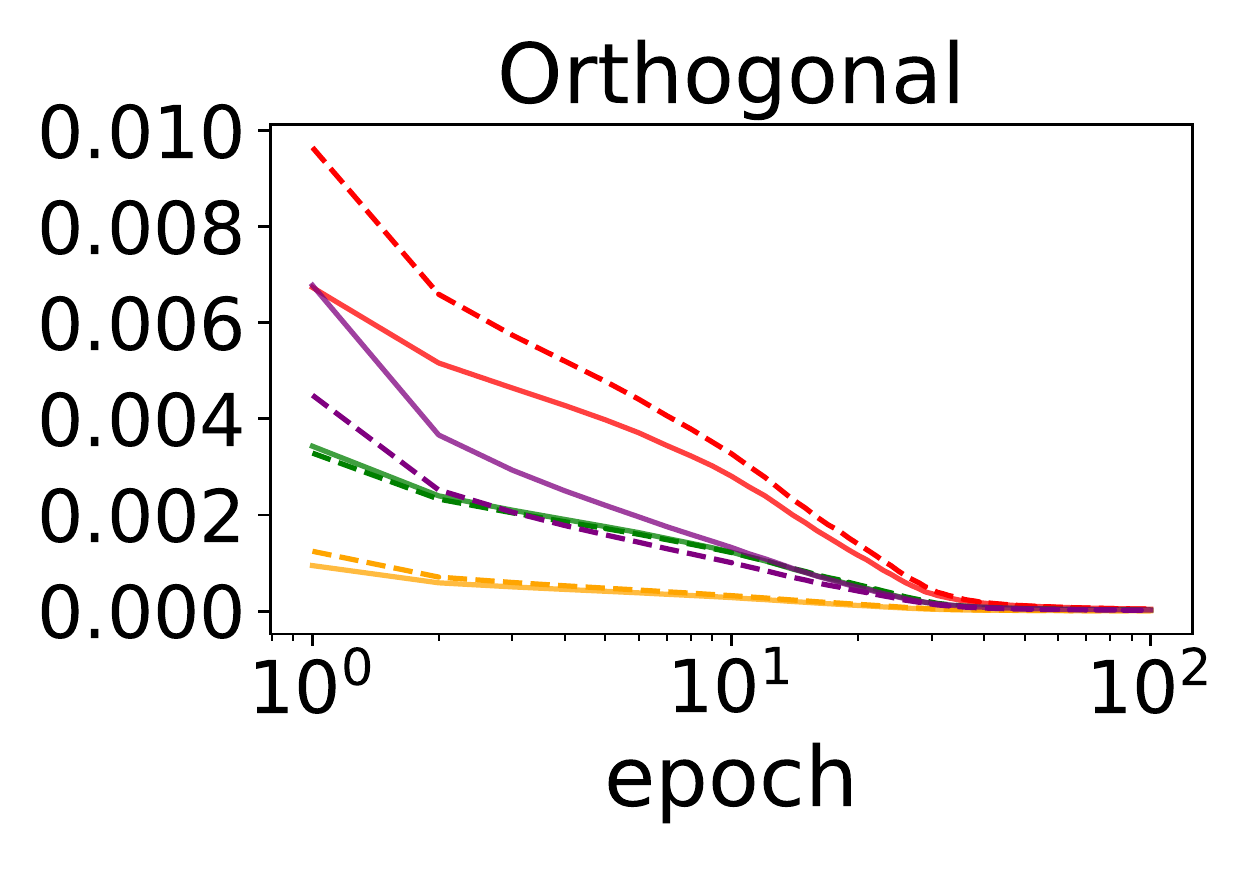}
     }
    
     \subfigure[CIFAR10]{
    \includegraphics[width=0.37 \linewidth]{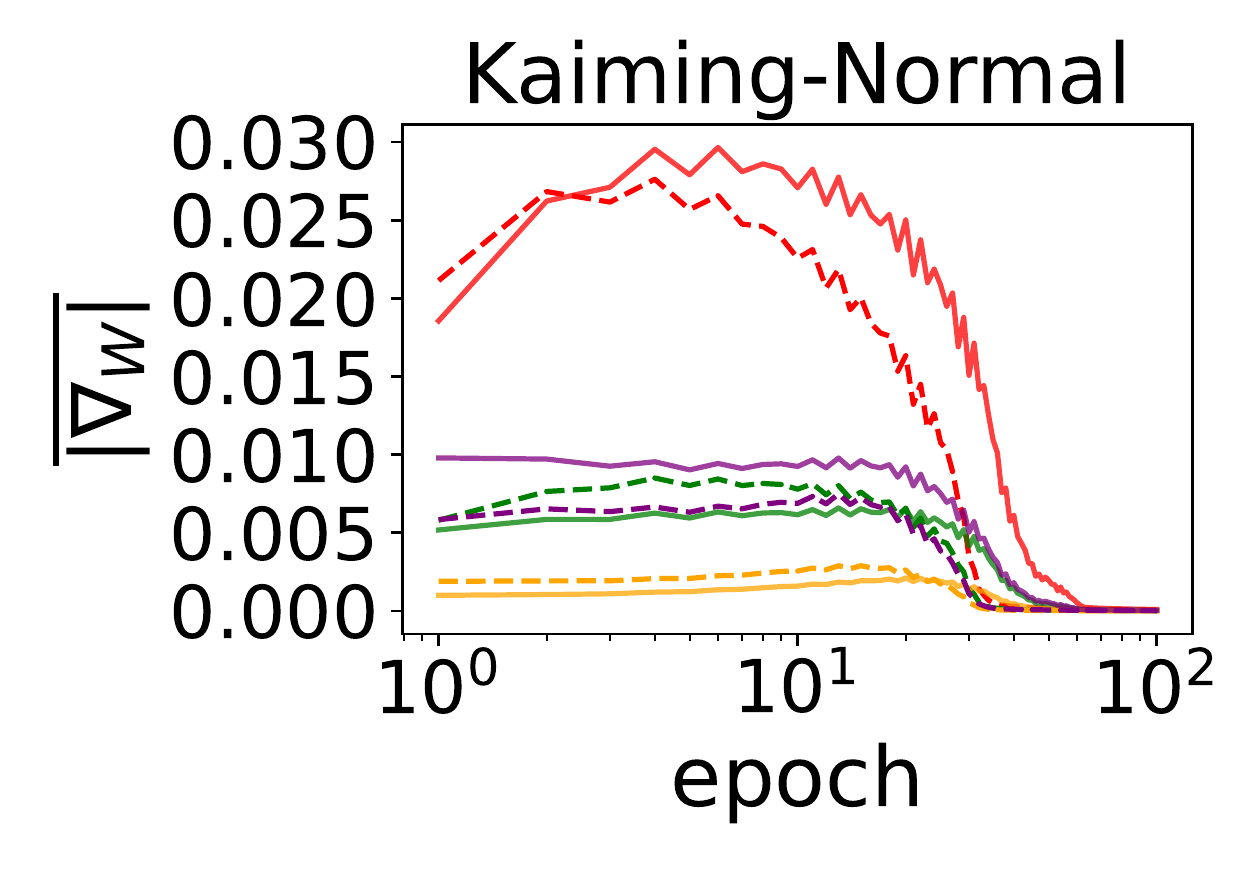} %\includegraphics[width=0.32 \linewidth]{gradients_legends.pdf}
    \hspace{2.75cm}
    \includegraphics[width=0.37 \linewidth]{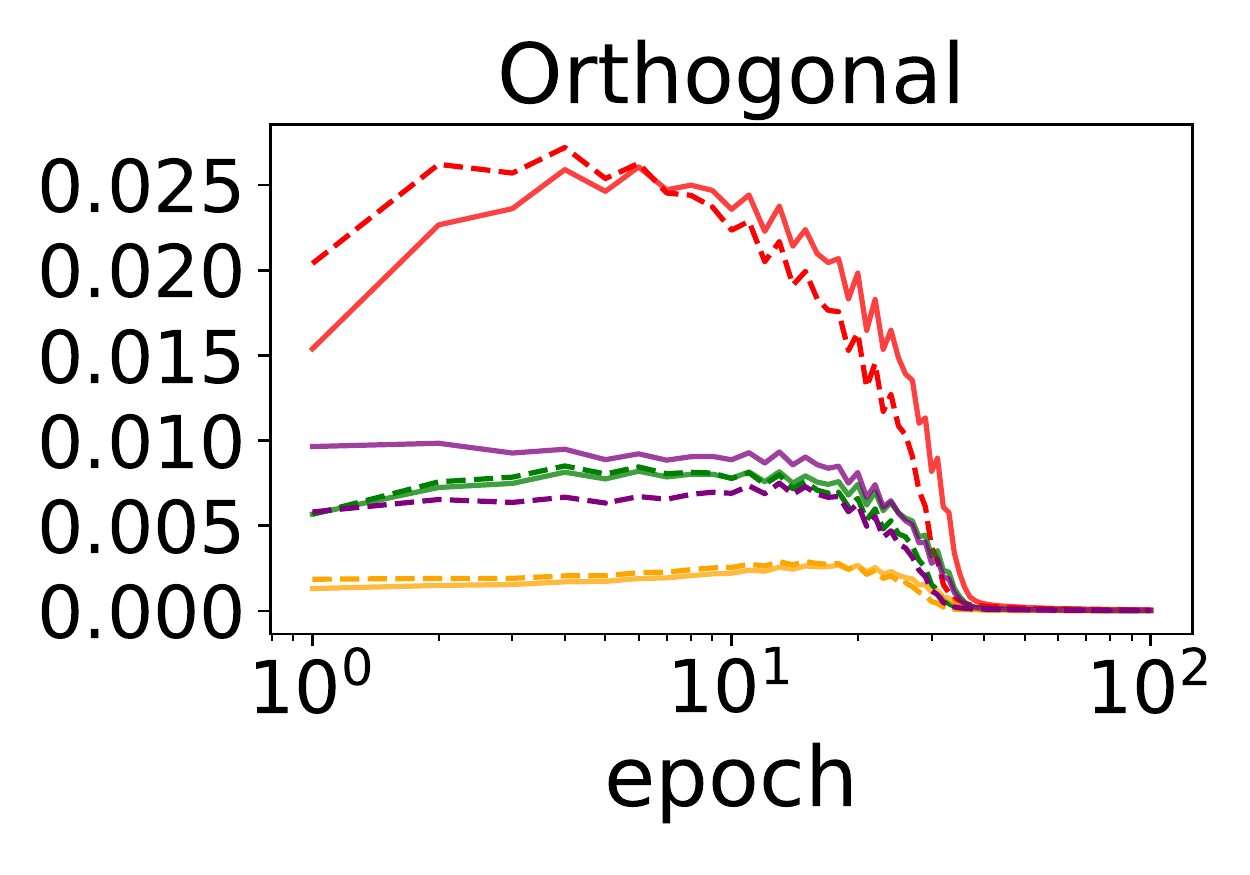} 
    }

    \caption{\label{fig:gradients}Average of the absolute gradient $\overline{|\nabla_W|}$ at each layer during the training of 100 LeNet networks on MNIST and CIFAR10, with two different weight initializers. The continuous line represents the behavior using the original weights, the dashed line is the effect after applying the proposed PA rewiring, and the colors represent the corresponding layer. The x-axis (epochs) is on a logarithmic scale for better visualization, since higher values appear at the first training epochs.}
\end{figure*}

To better understand the effects of PA rewiring during training, we compute the average absolute gradient flow $\overline{|\nabla_W|}$ at each epoch, for each layer of the LeNet models, and show the results in Figure~\ref{fig:gradients}. Each curve represents the average over the 100 models, with continuous lines related to the original weights and dashed lines for the models using PA rewiring. One can notice that the gradient flow is reduced at the output layer, and this behavior is consistent using both Kaiming-normal and Orthogonal weights, for both datasets. This effect is related to the lower classification loss achieved by the models using PA rewiring. On the other hand, the gradient flow increases in the preceding layers (at the initial epochs) and the models converge faster. This is an important behavior that helps to avoid the vanishing gradient problem. Therefore, the fact that higher gradient values propagate to preceding layers even when the classification loss is lower, corroborates that PA rewiring improves the backward signal flow during training.

\subsection{Systematic comparison}

A broad empirical evaluation is performed to assess the impact of the proposed PA rewiring method compared to literature weight initializers in different scenarios. Table~\ref{tab:comparisons} shows the results obtained with the LeNet architecture for different weight initializers on all three datasets (MNIST, FMNIST, and CIFAR10). Significant improvements in training performance are observed in the majority of the cases, while some isolated cases perform similarly to the base weights. Significant test accuracy improvements are also observed for the majority of the cases, except for Orthogonal weights on FMNIST, the only case with a measurable decrease in performance after applying PA rewiring. We also evaluated the proposed method  
using a small and a deep FC architecture under the same setup, and similar effects were observed.

\begin{table*}[!htb]
	\centering
	\caption{\label{tab:comparisons}Comparison of the proposed PA rewiring method applied on different weight initializers when using the LeNet architecture, trained on the MNIST, FMNIST, and CIFAR10 datasets.}
	\begin{tabular}{cc|ccc|cc}
		\hline
    % 	&&\multicolumn{3}{c|}{convergence (epochs)} &\multicolumn{3}{c}{test accuracy ($\%$)} \\
    % 	& method &min&average&max&min&average& max \\
		& & \multicolumn{3}{c|}{training metrics} & \multicolumn{2}{c}{test acc.}  \\
		
		dataset&weights& ep.\ 1 acc.\ &ep.\ 1 val.\ acc.\ &convergence&average&median\\ 
		\hline \hline
\multirow{8}{*}{\rotatebox[origin=c]{90}{MNIST}}
& Kaiming-uniform&96.25{\tiny$\pm$1.44}&95.86{\tiny$\pm$1.42}&15.56{\tiny$\pm$6.55}&99.06{\tiny$\pm$0.12}&99.06{\tiny$\pm$0.12} \\
& PA rewiring&\cellcolor{green!45}97.36{\tiny$\pm$0.31}&\cellcolor{green!45}96.84{\tiny$\pm$0.35}&\cellcolor{green!45}10.54{\tiny$\pm$2.25}&\cellcolor{green!45}99.10{\tiny$\pm$0.07}&\cellcolor{green!45}99.10{\tiny$\pm$0.05} \\
& Kaiming-normal&95.93{\tiny$\pm$2.31}&95.54{\tiny$\pm$2.26}&16.10{\tiny$\pm$6.59}&99.01{\tiny$\pm$0.13}&99.03{\tiny$\pm$0.13} \\
& PA rewiring&\cellcolor{green!45}97.37{\tiny$\pm$0.31}&\cellcolor{green!45}96.85{\tiny$\pm$0.32}&\cellcolor{green!45}10.22{\tiny$\pm$2.40}&\cellcolor{green!45}99.09{\tiny$\pm$0.07}&\cellcolor{green!45}99.09{\tiny$\pm$0.05} \\
& truncated-normal&96.18{\tiny$\pm$1.67}&95.81{\tiny$\pm$1.65}&15.77{\tiny$\pm$8.30}&99.05{\tiny$\pm$0.17}&99.08{\tiny$\pm$0.17} \\
& PA rewiring&\cellcolor{green!45}97.34{\tiny$\pm$0.32}&\cellcolor{green!45}96.81{\tiny$\pm$0.33}&\cellcolor{green!45}10.52{\tiny$\pm$2.37}&\cellcolor{green!45}99.09{\tiny$\pm$0.07}&\cellcolor{gray!65}99.09{\tiny$\pm$0.04} \\
& Orthogonal&97.08{\tiny$\pm$0.41}&96.74{\tiny$\pm$0.44}&12.40{\tiny$\pm$2.37}&99.19{\tiny$\pm$0.06}&99.19{\tiny$\pm$0.06} \\
& PA rewiring&\cellcolor{gray!65}97.13{\tiny$\pm$0.40}&\cellcolor{gray!65}96.72{\tiny$\pm$0.42}&\cellcolor{green!45}11.63{\tiny$\pm$2.28}&\cellcolor{gray!65}99.19{\tiny$\pm$0.06}&\cellcolor{gray!65}99.18{\tiny$\pm$0.03} \\
\hline
\multirow{8}{*}{\rotatebox[origin=c]{90}{FMNIST}}
& Kaiming-uniform&84.57{\tiny$\pm$0.92}&84.01{\tiny$\pm$0.85}&12.34{\tiny$\pm$1.26}&90.54{\tiny$\pm$0.23}&90.52{\tiny$\pm$0.23} \\
& PA rewiring&\cellcolor{green!45}85.62{\tiny$\pm$0.85}&\cellcolor{green!45}84.96{\tiny$\pm$0.81}&\cellcolor{green!45}10.91{\tiny$\pm$1.24}&\cellcolor{green!45}90.63{\tiny$\pm$0.22}&\cellcolor{green!45}90.64{\tiny$\pm$0.16} \\
& Kaiming-normal&84.45{\tiny$\pm$0.89}&83.87{\tiny$\pm$0.85}&12.77{\tiny$\pm$1.74}&90.55{\tiny$\pm$0.23}&90.52{\tiny$\pm$0.23} \\
& PA rewiring&\cellcolor{green!45}85.56{\tiny$\pm$0.86}&\cellcolor{green!45}84.82{\tiny$\pm$0.81}&\cellcolor{green!45}10.91{\tiny$\pm$1.32}&\cellcolor{green!45}90.64{\tiny$\pm$0.19}&\cellcolor{green!45}90.61{\tiny$\pm$0.13} \\
& truncated-normal&84.50{\tiny$\pm$1.04}&83.95{\tiny$\pm$0.97}&12.64{\tiny$\pm$1.64}&90.53{\tiny$\pm$0.24}&90.54{\tiny$\pm$0.24} \\
& PA rewiring&\cellcolor{green!45}85.62{\tiny$\pm$0.87}&\cellcolor{green!45}84.95{\tiny$\pm$0.87}&\cellcolor{green!45}11.11{\tiny$\pm$1.22}&\cellcolor{green!45}90.66{\tiny$\pm$0.20}&\cellcolor{green!45}90.66{\tiny$\pm$0.14} \\
& Orthogonal&83.61{\tiny$\pm$1.04}&83.23{\tiny$\pm$1.03}&13.05{\tiny$\pm$1.06}&91.06{\tiny$\pm$0.19}&91.06{\tiny$\pm$0.19} \\
& PA rewiring&\cellcolor{green!45}83.98{\tiny$\pm$1.10}&\cellcolor{green!45}83.58{\tiny$\pm$1.11}&\cellcolor{gray!65}12.94{\tiny$\pm$1.34}&\cellcolor{red!45}90.98{\tiny$\pm$0.20}&\cellcolor{red!45}90.97{\tiny$\pm$0.11} \\
\hline
\multirow{8}{*}{\rotatebox[origin=c]{90}{CIFAR10}}
& Kaiming-uniform&49.91{\tiny$\pm$1.67}&48.88{\tiny$\pm$1.50}&11.09{\tiny$\pm$2.04}&67.65{\tiny$\pm$0.61}&67.69{\tiny$\pm$0.61} \\
& PA rewiring&\cellcolor{green!45}52.87{\tiny$\pm$0.47}&\cellcolor{green!45}51.29{\tiny$\pm$0.53}&\cellcolor{green!45}9.74{\tiny$\pm$0.67}&\cellcolor{green!45}68.00{\tiny$\pm$0.56}&\cellcolor{green!45}67.97{\tiny$\pm$0.43} \\
& Kaiming-normal&50.17{\tiny$\pm$1.60}&49.09{\tiny$\pm$1.34}&10.95{\tiny$\pm$1.98}&67.59{\tiny$\pm$0.65}&67.57{\tiny$\pm$0.65} \\
& PA rewiring&\cellcolor{green!45}53.30{\tiny$\pm$0.56}&\cellcolor{green!45}51.57{\tiny$\pm$0.58}&\cellcolor{green!45}9.60{\tiny$\pm$0.80}&\cellcolor{green!45}68.03{\tiny$\pm$0.53}&\cellcolor{green!45}68.07{\tiny$\pm$0.37} \\
& truncated-normal&50.10{\tiny$\pm$2.07}&49.06{\tiny$\pm$1.85}&10.86{\tiny$\pm$1.97}&67.68{\tiny$\pm$0.64}&67.68{\tiny$\pm$0.64} \\
& PA rewiring&\cellcolor{green!45}52.83{\tiny$\pm$0.50}&\cellcolor{green!45}51.32{\tiny$\pm$0.62}&\cellcolor{green!45}9.64{\tiny$\pm$0.77}&\cellcolor{green!45}67.90{\tiny$\pm$0.54}&\cellcolor{green!45}67.98{\tiny$\pm$0.40} \\
& Orthogonal&50.65{\tiny$\pm$0.54}&49.84{\tiny$\pm$0.67}&9.98{\tiny$\pm$0.20}&69.08{\tiny$\pm$0.48}&69.08{\tiny$\pm$0.48} \\
& PA rewiring&\cellcolor{green!45}51.39{\tiny$\pm$0.50}&\cellcolor{green!45}50.42{\tiny$\pm$0.53}&\cellcolor{gray!65}9.98{\tiny$\pm$0.20}&\cellcolor{green!45}69.24{\tiny$\pm$0.49}&\cellcolor{green!45}69.26{\tiny$\pm$0.34} \\
\hline
      
	\end{tabular}
\end{table*}

\subsection{Deeper Models and Advanced Learning Schedules}\label{sec:deeper}

To measure the impact of the proposed PA rewiring method in a more complex scenario, we consider deeper models and more rigorous training schedules. In some cases, we adapted the training schedule to be more suitable to the available computing resources (balancing batch size with VRAM usage and time to train), and to achieve convergence on CIFAR10 (also considering that we split off an additional validation set, resulting in less training samples). The following architectures are analyzed with the CIFAR10 dataset:
\begin{itemize}
    \item ResNet (18 and 50)~\cite{he2016deep}: For ResNet18, the training configuration is the same as in the previous section. For ResNet50, training is done using the AdamW optimizer~\cite{loshchilov2017decoupled} for 150 epochs with cosine learning rate schedule with an initial learning rate of 0.02 (annealed to 0, no restarting), weight decay of 0.005, batch size 512, and gradient norm clipping.
    
    \item  SL-ViT \cite{lee2021vision}: Transformer architectures have recently demonstrated great potential for vision tasks~\cite{dosovitskiy2020image} (ViT). The SL-ViT is a variant of the original ViT that is more suitable for training from scratch with smaller datasets. The method is trained as in~\cite{lee2021vision} \footnote{\url{www.github.com/aanna0701/SPT_LSA_ViT}}: using the AdamW optimizer, 100 epochs, cosine learning rate schedule with an initial learning rate of 0.003 (annealed to 0, no restarting), batch size 128, 10 warmup epochs, and weight decay of 0.05. For regularization, we use label smoothing~\cite{szegedy2016rethinking}, stochastic depth~\cite{huang2016deep}, CutMix~\cite{yun2019cutmix}, and Mixup~\cite{zhang2017mixup}.
    
    % \item SL-Swin \cite{lee2021vision}: SL-Swin is a variant of the Swin architecture \cite{liu2021swin}, an improved vision transformer if compared to the previous ViT. The SL variant is more suitable for smaller-scale datasets. The method is trained as in \cite{lee2021vision}, i.e., in the same way as for SL-ViT.

    \item ConvMixer \cite{trockman2022patches}: A more recent architecture that mixes the spatial and channel locations of patch embeddings using only standard convolutions. We considered patch size 2, kernel size 5, hidden dimension 256, and depth 8. The model is trained similarly as in~\cite{trockman2022patches} \footnote{\url{www.github.com/locuslab/convmixer-cifar10}}, considering 100 epochs with the AdamW optimizer, triangular learning rate scheduling with a maximum learning rate of 0.1, weight decay of 0.005, batch size 512, and gradient norm clipping.
\end{itemize} 

For the training of all deep models (except ResNet18), the following data augmentation is employed in addition to the procedures already described above: random scaling, RandAugment~\cite{cubuk2020randaugment}, random erasing~\cite{zhong2020random}, and color jitter. We keep the same configuration of random seeds as previously described: each repetition changes Pytorch's manual seed responsible for random weight sampling, while the training/validation split is kept, and PA rewiring is applied to the weights generated with the same seeds. We performed the statistical tests for the ResNet18 experiments, since we performed 100 repetitions to each weight initializer with this architecture, while 10 repetitions were done for the other models due to computing constraints. This is done since we are using much bigger models and strong data augmentation, making it unfeasible to store all data in GPU memory for faster computation, and batch generation needs to be done in the CPU. In this scenario, the differences in training time on a 32-core CPU with an RTX 2080 ti GPU were around 30 to 60 times compared to the previous experiments with smaller models without data augmentation. Note again that we keep our three data splits policy and experimental constraints to achieve a more robust evaluation of the impacts of weights, rather than to perform a search for the best possible performance (e.g., state-of-the-art results on CIFAR10~\cite{wang2021sample} may require more than 600 training epochs). Nonetheless, since we isolate all other stochastic and random procedures that could interfere with performance, we believe that 10 repetitions gives a good estimate of the impact of the weights alone. Note, however, that we do not perform statistical tests in this case since the sample size is too small. In addition, we also compare the maximum obtained test accuracy of each model.

The results for the experiments with the deeper models are shown in Table~\ref{tab:comparisons_deeper}. As base weights, we considered the Kaiming-normal initialization for ResNets, which is the proposed initialization for this specific architecture, and Orthogonal weights for the other models, since it achieved the best results in our previous experiments. For ResNet18, the proposed PA rewiring method increases the training performance, speeds up convergence, and reaches higher test results at the end, all with statistically significant differences. Note that since no data augmentation is used in this case, the lowest validation loss is reached with around seven epochs only (with a small standard deviation). This does not seem to affect the gains after training, where rewiring improved the average performance of the models and also the highest obtained test accuracy. For the other cases, the method also provides similar effects, improving average performance and the maximum obtained test accuracy. However, the convergence time in this scenario is higher when using any given initial weights, highlighting the effects of strong regularization. In other words, although PA rewiring increases the performance at the beginning of training, this does not mean that models will train faster in this case. On the other hand, it still provides performance improvements after training in the majority of cases, implying a better generalization power. Although a limited number of trials are performed in this case, we believe that the results corroborate the previous findings with both the LeNet and ResNet18 architectures.

\begin{table*}[!htb]
	\centering
	\caption{\label{tab:comparisons_deeper}Comparison of weight initializations using deeper architectures and more complex training schedules on the CIFAR10 dataset. ResNet18* was trained according to the simpler schedule (SGD, 100 epochs), and 100 repetitions were conducted (along with statistical tests).}
	
	\begin{tabular}{cc|ccc|ccc}
		\hline

		& & \multicolumn{3}{c|}{training metrics} & \multicolumn{3}{c}{test acc.}  \\
		
		model&weights& ep.\ 1 acc.& ep.\ 1 val.\ acc.\ &convergence& average&median&max.\\ 
		\hline \hline

        \multirow{2}{*}{\rotatebox[origin=c]{0}{ResNet18*}}
& Kaiming-normal&63.48{\tiny$\pm$0.64}&57.85{\tiny$\pm$0.62}&7.48{\tiny$\pm$0.88}&72.94{\tiny$\pm$0.53}&72.94{\tiny$\pm$0.53}&74.23 \\
& PA rewiring&\cellcolor{green!45}64.71{\tiny$\pm$0.60}&\cellcolor{green!45}58.61{\tiny$\pm$0.64}&\cellcolor{green!45}7.08{\tiny$\pm$1.00}&\cellcolor{green!45}73.36{\tiny$\pm$0.49}&\cellcolor{green!45}73.37{\tiny$\pm$0.36}&74.36 \\
% & Orthogonal&64.92{\tiny$\pm$0.56}&60.35{\tiny$\pm$0.61}&7.28{\tiny$\pm$0.96}&74.51{\tiny$\pm$0.44}&74.48{\tiny$\pm$0.44}&75.50 \\
% & PA rewiring&\cellcolor{green!25}65.48{\tiny$\pm$0.60}&\cellcolor{green!25}60.84{\tiny$\pm$0.64}&\cellcolor{gray!25}7.04{\tiny$\pm$1.00}&\cellcolor{gray!25}74.60{\tiny$\pm$0.48}&\cellcolor{gray!25}74.67{\tiny$\pm$0.36}&\textbf{75.61} \\
\hline

\multirow{2}{*}{\rotatebox[origin=c]{0}{ResNet50}}
& Kaiming-normal&13.93{\tiny$\pm$3.24}&13.92{\tiny$\pm$3.17}&141.10{\tiny$\pm$5.05}&92.29{\tiny$\pm$0.35}&92.28{\tiny$\pm$0.35}&92.74 \\
& PA rewiring&14.06{\tiny$\pm$2.74}&13.91{\tiny$\pm$2.61}&139.00{\tiny$\pm$5.71}&92.37{\tiny$\pm$0.26}&92.35{\tiny$\pm$0.26}&92.75 \\
% & Orthogonal&18.17{\tiny$\pm$2.42}&18.16{\tiny$\pm$2.46}&141.20{\tiny$\pm$5.71}&92.09{\tiny$\pm$0.34}&92.12{\tiny$\pm$0.34}&92.62 \\
% & PA rewiring&15.84{\tiny$\pm$3.53}&15.89{\tiny$\pm$3.57}&136.50{\tiny$\pm$5.63}&92.14{\tiny$\pm$0.16}&92.10{\tiny$\pm$0.16}&92.41 \\

\hline
\multirow{2}{*}{\rotatebox[origin=c]{0}{SL-ViT}}
& Orthogonal&22.64{\tiny$\pm$0.61}&22.42{\tiny$\pm$0.68}&108.00{\tiny$\pm$0.00}&89.10{\tiny$\pm$0.29}&89.20{\tiny$\pm$0.29}&89.52 \\
& PA rewiring&24.43{\tiny$\pm$0.94}&24.35{\tiny$\pm$0.91}&108.00{\tiny$\pm$0.00}&89.13{\tiny$\pm$0.26}&89.06{\tiny$\pm$0.26}&89.57 \\
\hline

% \multirow{2}{*}{\rotatebox[origin=c]{0}{SL-Swin}}
% & Orthogonal&25.20{\tiny$\pm$0.88}&25.62{\tiny$\pm$1.11}&107.70{\tiny$\pm$0.90}&93.71{\tiny$\pm$0.19}&93.73{\tiny$\pm$0.19}&94.02\\
% & PA rewiring&23.82{\tiny$\pm$0.92}&24.01{\tiny$\pm$0.96}&108.00{\tiny$\pm$0.00}&93.75{\tiny$\pm$0.20}&\textbf{93.83}{\tiny$\pm$0.20}&94.00 \\
% \hline

\multirow{2}{*}{\rotatebox[origin=c]{0}{ConvMixer}}
& Orthogonal&39.18{\tiny$\pm$1.30}&38.54{\tiny$\pm$1.35}&95.70{\tiny$\pm$1.79}&93.69{\tiny$\pm$0.14}&93.70{\tiny$\pm$0.14}&93.84 \\
& PA rewiring&39.91{\tiny$\pm$1.26}&39.16{\tiny$\pm$1.31}&95.50{\tiny$\pm$2.46}&\textbf{93.77}{\tiny$\pm$0.13}&\textbf{93.78}{\tiny$\pm$0.13}&\textbf{93.95} \\

\hline

	\end{tabular}
\end{table*}

\subsection{Discussion}

Our results corroborate previous works~\cite{testolin2020deep,zambra2020emergence,you2020graph,scabini2021structure,la2021characterizing} on the complex network properties of ANNs, reinforcing the existence of these properties and their importance. Random initialization was a key factor in most of the mentioned works, as discussed by most authors. For instance, the only degree of freedom causing the correlation between structure and performance observed in~\cite{scabini2021structure} was the random weight organization (the training procedure was fixed and a single initializer was used). Our observations on the strength variance of randomly initialized neurons partially explain this phenomenon, corroborating that the performance differences are caused also by the weight organization at initialization, rather than by the training procedure alone. This is also corroborated by correlating our results with the theoretical findings of~\cite{huang2021quantifying} and the empirical result of~\cite{picard2021torch,wightman2021resnet} (on the variance caused by random seeds). In this sense, we observe a similar behavior as these previous works by varying only the weight initialization seed, suggesting that random weights are a key factor in causing such variance.

%In conclusion, these findings highlight the importance of random initialization and the possibilities of ANN improvements through Network Science.

%It is also relevant to mention the findings of \cite{jesus2021effect} and the LTH, which corroborates some effects caused by specific weight initializations. Considering the proposed PA rewiring method, the obtained parameter space after rewiring seems to be one of such special cases

%The results achieved through PA rewiring corroborate the findings of \cite{jesus2021effect}, which states that ANNs that are "strong learners" converges to the neighborhood of their initial configuration. we argue that the parameter space obtained by reorganizing weights with the PA rewiring method provides.
% It is also worth mentioning that neuronal strength also resembles random walk initialization \cite{sussillo2014random}, as walks tend to be attracted by hubs. 

Empirical results show that the proposed PA rewiring method causes significant effects on ANN performance. More specifically, performance is considerably improved after a single training epoch, corroborating to better initial 
% make sure to check, there was a strange 'begging'
% maybe: "More specifically, performance improves at initial training, ..."? 
parameter space, and the convergence time is decreased in most cases. More surprisingly, it also improves the test performance of final models, leading to a better generalization power. Another effect noticed in most experiments is the decrease in performance variance, considering both the standard deviation and the median absolute deviation. However, it is important to notice that in some specific cases it may decrease the final performance of the models, as results with a shallow architecture and simple training schedule show. We believe that, in this case, the performance gains at the beginning of training cause models to overfit too fast (see convergence time), thus reaching poor local minima that a simple training schedule is not able to overcome. Therefore, this effect may be also related to the training schedule and not only to the initial weights, since better training seems to overcome this limitation. For instance, when more complex training schedules with strong regularization are employed, the models achieve both higher training and test performance in general. The effects of PA rewiring in this latter case, which is the common scenario in state-of-the-art research with large-scale vision models, highlights the importance of the initial random weights and their organization.

Regarding the combination of PA rewiring with other deep learning techniques, our results show that it works well in conjunction with batch normalization~\cite{ioffe2015batch} and residual connections~\cite{he2016deep}, two of the most important modules of state-of-the-art ANNs. It also works well with more complex training schedules, and with more recent architectures, such as ViT and ConvMixer. Regarding other kinds of weight operations, such as weight normalization~\cite{salimans2016weight} and other methods that make the learned function invariant to scaling of the weights, the proposed PA rewiring goes in a different direction as we do not change the scale and shape of the weight distribution. In other words, as our approach only reorganizes weights at initialization, it can be used together with any of these methods. Analyzing the impacts of combining our approach with all such methods is beyond the scope of this work, as there is a large number of possibilities. Nevertheless, we believe that the selected methods, architectures, datasets, and the statistical analysis employed in this work give a good estimate of the behavior of PA rewiring in practice. We also avoided doing analyses of individual models and neurons as this would be highly influenced by the randomness of the initializers. For instance, we believe that these specific analyses, which are common in the ANN literature, are one of the potential flaws when dealing with several degrees of freedom, causing some findings to be hard to replicate and generalize. Therefore, our analyses focused on distributions and patterns observed collectively on a wide range of random networks, under different scenarios and training dynamics.

\section{Conclusion}\label{sec:conclusion}

A new method to improve weight initialization is proposed for general ANN architectures. To the best of our knowledge, this is the first work to directly explore Network Science concepts in the context of random weight initialization of deep ANNs. First, we showed a measurable correlation with statistical confidence between the neuronal strength at random initialization and the final performance of a LeNet model. We then explored neuronal strength properties to improve initialization, showing that a higher strength variance decreases performance, while a lower variance increases it. These concepts were considered to develop a new heuristic, called PA rewiring, which works in conjunction with any weight initialization method by reorganizing neuronal connections in a way that their strength distributions tend to zero. In other words, we modify the long-tailed strength distribution produced by literature methods, and the results show that this simple procedure improves training and test performance. The method works locally at each ANN layer, without external knowledge from the architecture or domain in which it is being employed, and has a relatively low computational budget since it is applied just once at initialization. 

Our empirical evaluation corroborates the importance of the random initialization of ANNs, even when deep models and complex training schedules are employed. The proposed PA rewiring method provides a simple and cheap way to optimize random weights, which can be easily coupled with state-of-the-art deep learning pipelines. Furthermore, we believe that the organization of the weights is an important aspect of random initialization that should be further investigated. Considering our findings, one interesting direction for future works is to explore other Network Science measures and tools. The proposed PA rewiring method can also be explored in different ways, for instance by using different levels of rewiring (instead of rewiring the whole layer), and/or applying it heterogeneously in different layers. Moreover, a new interesting approach would be to effectively generate random values based on the network structural properties, rather than relying on literature initializers. Another possibility is to explore the method in different scenarios. Although we consider only compute vision architectures in our experiments, the proposed method is based on generic weight matrices. In this sense, it can be simply coupled with existing initialization policies for different architectures and in different domains.

\section*{Acknowledgments}
L. Scabini acknowledges funding from the São Paulo Research Foundation (FAPESP) (Grants \#2019/07811-0 and \#2021/09163-6) and the National Council for Scientific and Technological Development (CNPq) (Grant \#142438/2018-9). O. Bruno acknowledges support from CNPq (grant \#307897/2018-4) and FAPESP (grants \#18/22214-6 and \#21/08325-2). The authors are also grateful to the NVIDIA GPU Grant Program for the donation of GPUs that were used in this research. This research received funding from the Flemish Government under the "Onderzoeksprogramma Artificiële Intelligentie (AI) Vlaanderen" programme.

\bibliographystyle{unsrt}
%\section*{\refname}
\bibliography{arxiv.bib}

\end{document}